\documentclass[10pt,twocolumn,letterpaper]{article}

\usepackage{iccv}
\usepackage{times}
\usepackage{epsfig}
\usepackage{graphicx}
\usepackage{amsmath}
\usepackage{amssymb}
\usepackage{subcaption}

\usepackage{authblk}
\usepackage{multirow}
\usepackage{graphicx}
\usepackage{booktabs}
\usepackage{multirow,bigstrut}
\usepackage{tabu}
\usepackage{bigstrut}

\newcommand{\hide}[1]{}

\usepackage[breaklinks=true,bookmarks=false]{hyperref}

\iccvfinalcopy 

\def\iccvPaperID{****} 

\begin{document}
\title{\emph{KeystoneDepth}: Visualizing History in 3D}
\makeatletter
\renewcommand\AB@affilsepx{, \protect\Affilfont}
\makeatother
\author[1]{Xuan Luo}
\author[1]{Yanmeng Kong}
\author[2]{Jason Lawrence}
\author[2]{Ricardo Martin-Brualla}
\author[1, 2]{Steve Seitz}
\affil[1]{University of Washington}
\affil[2]{Google}

\twocolumn[{%
\renewcommand\twocolumn[1][]{#1}%
\maketitle

\begin{center}
\vspace*{-10pt}
\setlength{\tabcolsep}{0.2mm}
\def\imh{68pt}
\begin{tabular}{cccc}
        \includegraphics[height=\imh]{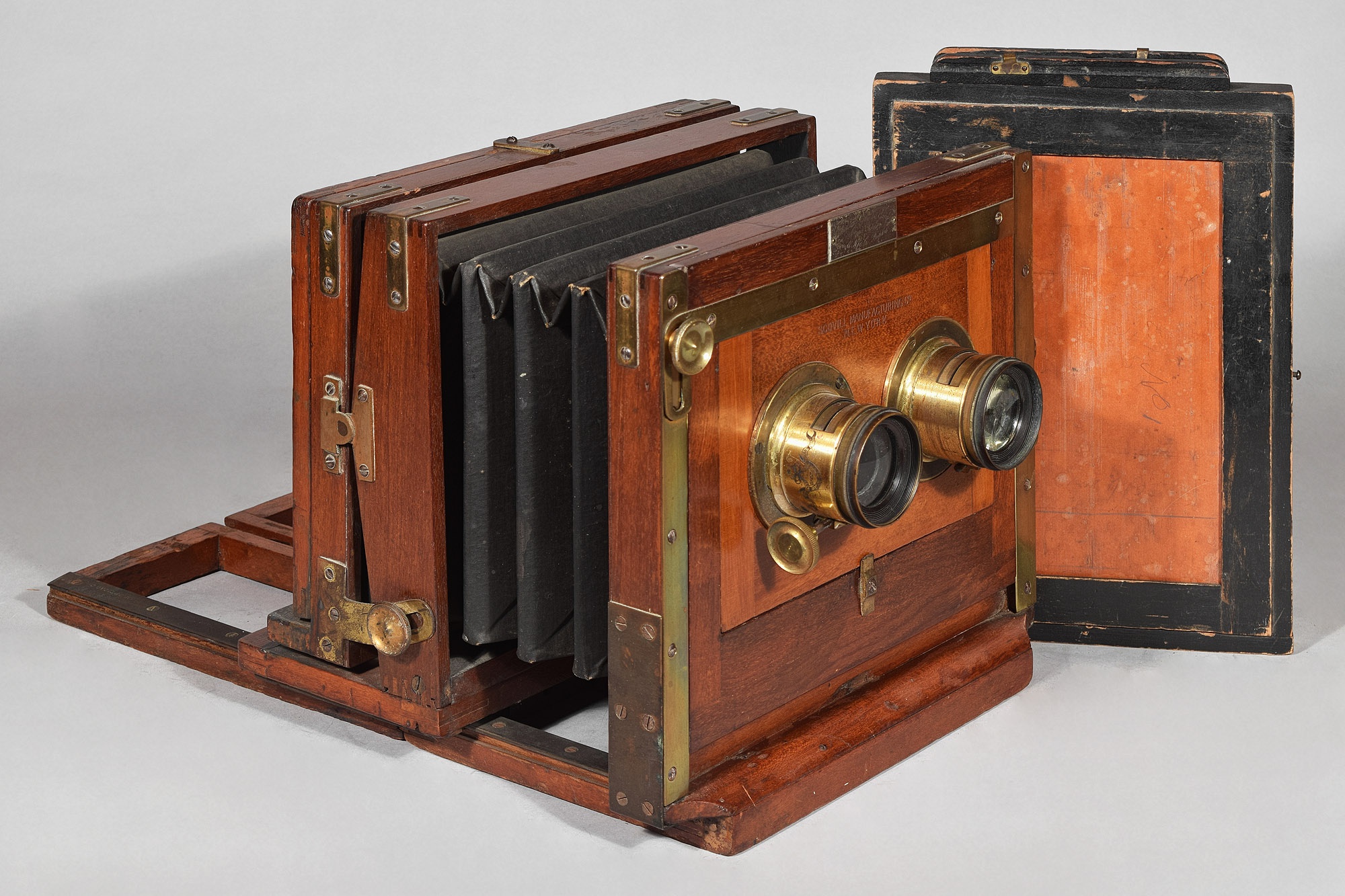} &
        \includegraphics[height=\imh]{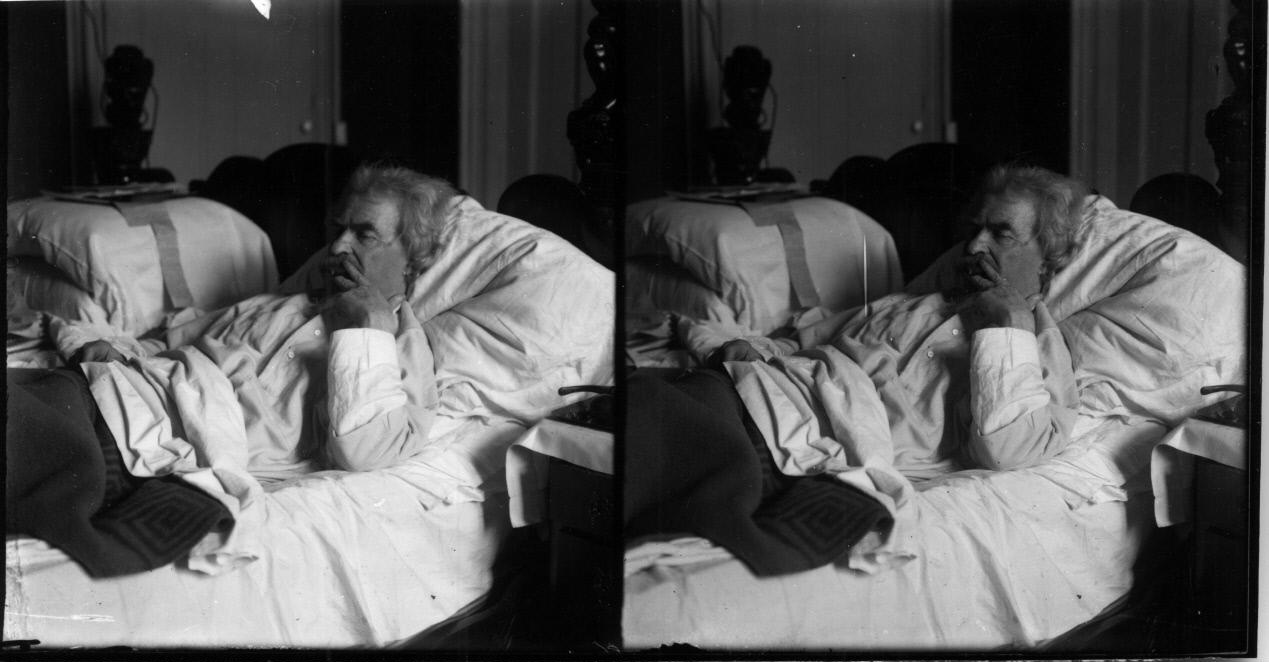} &
        \includegraphics[height=\imh]{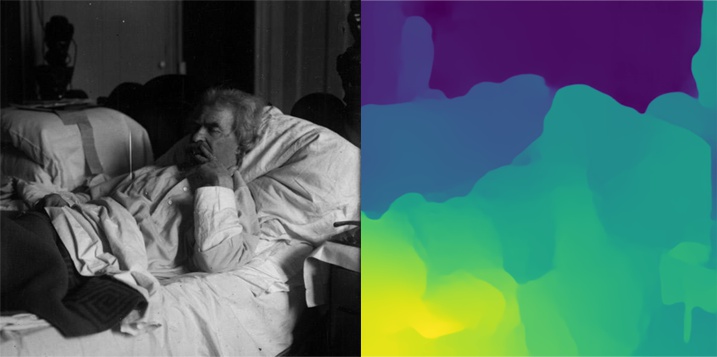} &
                \includegraphics[height=\imh]{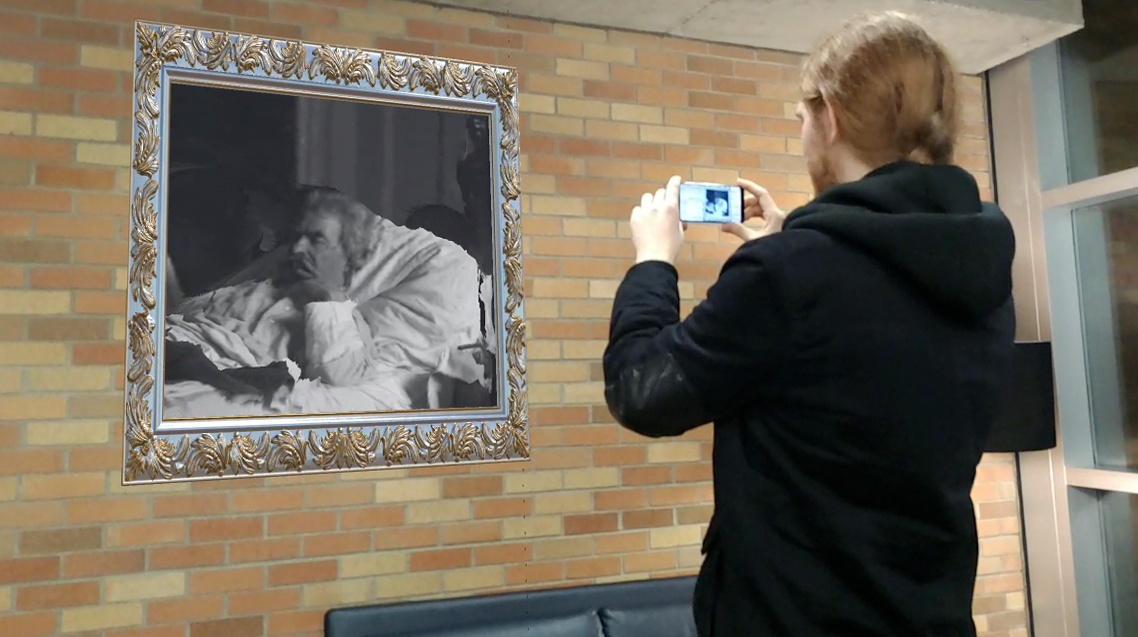}\\
\small{(a) Antique Stereo Camera} & \small{(b) Antique Stereograph} & \small{(c) \emph{KeystoneDepth} Collection} & \small{(d) \emph{KeystoneAR} App}
\end{tabular}
\vspace*{5pt}
    \captionof{figure}{We introduce {\em KeystoneDepth}, a collection of over 10,000 rectified antique stereographs of historical scenes captured between 1860 and 1963.
    (a) Stereo cameras were invented in the 1850s and (b) hundreds of thousands of antique stereographs are available today, like this one of Mark Twain lying in bed. (c) We applied multiple processing steps to produce clean stereo image pairs, complete with calibration data, rectification transforms, and depthmaps. (d) We also introduce a novel lightweight view synthesis technique that powers a mobile AR application, {\em KeystoneAR}, that gives the sensation of looking through an open window onto these historical scenes. Please see our supplementary material for many more results.
\label{fig:teaser}}
\end{center}%
}]

\begin{abstract}
   This paper introduces the largest and most diverse collection of rectified stereo image pairs to the research community, KeystoneDepth, consisting of tens of thousands of stereographs of historical people, events, objects, and scenes between 1860 and 1963.  Leveraging the Keystone-Mast raw scans from the California Museum of Photography, we apply multiple processing steps to produce clean stereo image pairs, complete with calibration data, rectification transforms, and depthmaps. A second contribution is a novel approach for view synthesis that runs at real-time rates on a mobile device, simulating the experience of looking through an open window into these historical scenes. We produce results for thousands of antique stereographs, capturing many important historical moments.
\end{abstract}
\section{Introduction}

Wouldn't it be fascinating to be in the same room as Abraham Lincoln, visit Thomas Edison in his laboratory, or step onto the streets of New York a hundred years ago? If only there were a way to travel back in time and capture these moments with a lightfield camera~\cite{overbeck18}! We take a major step towards this lofty goal by publishing the \emph{KeystoneDepth} collection, the world's largest and most diverse collection of stereo image pairs, showing historic people, events, objects, and scenes between 1860 and 1963. We also describe a novel real-time view synthesis technique we developed for visualizing this collection that reproduces stereo, parallax, and visibility cues over a predetermined head volume to give the sensation of peering through a window into these other times and places.

Stereo cameras and viewers were invented in the mid 1850s, and quickly became very popular. Many of the world’s most famous people, events, and places over the following century were captured in stereo, and hundreds of thousands of these stereographs survive to this day.  Notably, the California Museum of Photography has 250,000 stereoscopic glass-plate and film negatives from the Keystone-Mast Collection, of which 45,197 are available online. This imagery opens up the fascinating possibility of reconstructing dense lightfields of these historic scenes that could be used to create immersive "like you were there" visualizations.

However, these antique stereographs are not in a form that facilitates this type of research and development. They are uncalibrated, unaligned, and contain many artifacts (scratches, damage, dirt, exposure differences, contrast loss, scanning errors, etc.). Modern view synthesis algorithms are not designed to work with this kind of data, and predictably do not produce satisfactory results when applied naively. Another major challenge is overcoming the need to extrapolate well beyond the input stereo pair in order to achieve an immersive range of head motion.


Our contributions are two-fold.  First, we introduce the {\em KeystoneDepth} collection, the world's largest collection of rectified antique stereographs with camera calibration, depth, and metadata, as a resource for the research community, see Section~\ref{sec:dataset}.  Beyond its historical value, this collection is one of the largest and most diverse collections of stereo images of any kind. Assembling, processing, and cleaning this collection took months of work, and we believe it is a contribution of significant value to the research community.

Second, we introduce a novel approach for view synthesis that is able to generate new views of a captured historic scene at real-time rates on a mobile device as the viewer moves within a predetermined head volume. Unlike traditional lightfield rendering techniques \cite{levoy96} that require a dense set of input images, our formulation requires only a single input stereo pair (similar to the method of Zhou et al.~\cite{zhou2018stereo}, but designed for 6DOF rather than 1DOF motion). In particular, we use a lightweight representation of the 3D scene comprised of several depth maps with aligned intensity images. We introduce a novel {\em double reprojection} method to train a deep neural network for inpainting intensity and depth values that are missing at interpolated viewpoints. One additional innovation is a depth boundary optimization technique that improves the quality of object silhouette edges under camera motion. Altogether, our approach can smoothly extrapolate views significantly beyond the input range at real-time rendering rates, thus creating a compelling illusion of looking through a window into one of these historic scenes.

\section{Related Work}

We describe related work in two different areas:  stereo datasets by prior authors, and techniques for producing 6DOF (parallax) enabled representations from stereo pairs.

\subsection{Stereo Datasets}

Several research groups have produced stereo image benchmark datasets over the years, most notably the Middlebury benchmark \cite{schar2002}, leading to rapid progress in stereo performance.  While the Middlebury dataset is limited in number (a few dozen) and scope (laboratory scenes), subsequent researchers \cite{Geiger2012CVPR,Knapitsch2017,Schops17} have improved the selection and quality of stereo benchmark images available online.
Nevertheless, these datasets are largely limited to buildings \cite{Schops17}, streets \cite{Geiger2012CVPR}, or specific objects like tanks and temples \cite{Knapitsch2017}.
While Li et al. \cite{MegaDepthLi18} use multiview stereo to infer depthmaps for tourist sites using Internet imagery, it's less clear how to produce rectified stereo pairs in this manner.


\subsection{Stereo to 6DOF}

Viewing a stereo image pair with a stereo viewer provides a compelling experience except that your viewpoint is fixed.  Simulating {\em parallax} as you move your head is key to a more realistic and immersive visualization.  The simplest form is {\em view interpolation} \cite{Chen93,Mahaj09,Seitz96}, which produces a narrow range of viewpoints strictly in between the two inputs.  While some view interpolation methods are capable of extrapolation, holes emerge and performance degrades beyond the input views due to visibility changes.

Filling in these holes, i.e., hallucinating hidden regions, is key to generating a larger (6DOF) range of viewpoints.
This is however very challenging, and previous work has typically resorted to using multiple cameras~\cite{chaurasia2013depth, hedman18, penner2017soft3dreconstruction, Zheng09,zitnick2004}. 
Zhou et al.~\cite{zhou2018stereo} proposed instead a {\em stereo magnification} approach that uses machine learning to estimate a multi-plane image representation from a single stereo view.
Although the estimation is highly unconstrained, they show high-quality view extrapolations results without holes in a variety of settings, that sometimes show depth quantization and rubber sheet artifacts at edges.
Recently, Choi et al.~\cite{choi2018extremenvs} generated extreme view extrapolations (up to 30x the baseline), but it requires running a CNN to render each  frame, making it unsuited for lightweight real-time applications.
Unlike~\cite{zhou2018stereo}, our proposed representation based on multiple inpainted intensity and depth images exhibits no depth quantization, and is lightweight to run in a cellphone.
To note as well, is the recently launched Facebook 3D Photo feature~\cite{Facebook3dPhotos}, that enables creating a 6-DOF experience from a stereo image taken on a cellphone, and uses  diffusion~\cite{Bertalmio2000ImageInpainting} to fill in missing regions.


\subsection{Image and Depth Inpainting}

Traditional image inpainting algorithms find candidate patches in the same image to fill in holes~\cite{barnes2009patchmatch, Bertalmio2000ImageInpainting, efros1999texture}, or find good matches in a large database~\cite{hays2007scene}. State-of-the-art methods~\cite{IizukaSIGGRAPH2017,partial-convolution} use large CNNs that do semantic aware inpainting by recognizing objects in the scene and learning how to inpaint them. Other work has addressed {\em depth} inpainting.  For example,
Zhang et al.~\cite{zhang2018deep} estimate scene normals and use them to inpaint depths with sparse constraints. Others~\cite{Holynski2018occlusion,shan2014occluding, valentin2018motionstereo} focused on densifying sparse depthmaps for 3d reconstruction and AR applications.
In this paper, we train networks to do both image and depth inpainting and build on the partial convolution framework of~\cite{partial-convolution}.
\section{The \emph{KeystoneDepth} Dataset} \label{sec:dataset}

A primary contribution of this work is the creation of a large public collection of rectified antique stereoscopic images along with corresponding camera parameters and disparity maps (scene depth up to an unknown scale). 
Hundreds of thousands of antique stereographs can be found in museums and antique stores, and many can be found online~\cite{NYPL, KeystoneMast, library-of-congress}.
Among them, the Keystone-Mast Collection~\cite{KeystoneMast} maintained by the California Museum of Photography (CMP) has the largest collection.
It mostly consists of gelatin silver contact prints made from the original negatives, captured by the Keystone View Company, a major distributor of stereoscopic images. 
This collection contains 250,000 stereographs captured between 1860 and 1963, portraying a wide variety of subjects, ranging from scenes from World War I and American presidential inaugurations to portraits of Mark Twain and Thomas Edison to images of the Egyptian pyramids and scenes of daily life on farms and factories during that time period. 
Of these, 45,197 are available online.

While the raw imagery is available, it is not in a form well-suited for research and development, as the images are not aligned, calibrated, or rectified.  
Furthermore, many of the scans are of poor quality or have significant visual artifacts (Figure~\ref{fig:bad-images}, top row).
A major challenge we faced in constructing this dataset was filtering out the many unusable samples from the Keystone-Mast collection and, for the rest, determining the largest artifact-free bounding box in each side of the stereo image that contained usable data, as described later in this section.

Each entry in the \emph{KeystoneDepth} dataset consists of the original stereoscopic image and its corresponding metadata, a URL to the corresponding entry in the Keystone-Mast collection, a bounding box for each image in the stereo pair marking the largest image region free enough of artifacts for reliable disparity estimation, a rectified stereo pair, camera parameters, and a pair of disparity maps. The resolution of the images is typically about $570\times 610$ pixels, and we provide depthmaps at $512\times 512$ after rescaling. Altogether, we considered 29,480 stereoscopic images from the Keystone-Mast collection and were able to compute disparity maps for 10,134. We plan to expand this further as more of the museum's holdings make their way online.

\subsection{Image Acquisition and Filtering}

\begin{figure}[t]
\centering
\setlength{\tabcolsep}{1mm}
\begin{tabular}{cc}
\includegraphics[width=0.48\columnwidth]{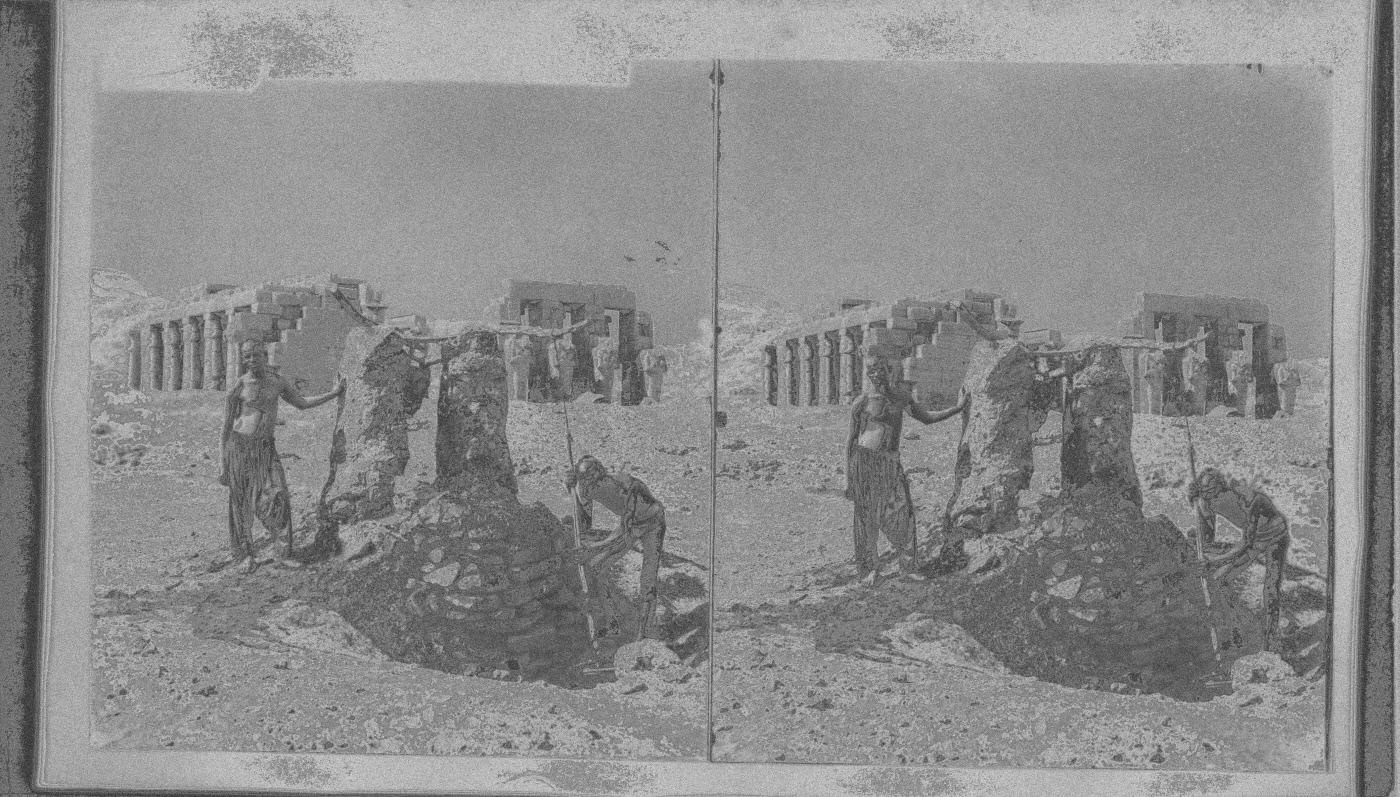} &  
\includegraphics[width=0.48\columnwidth]{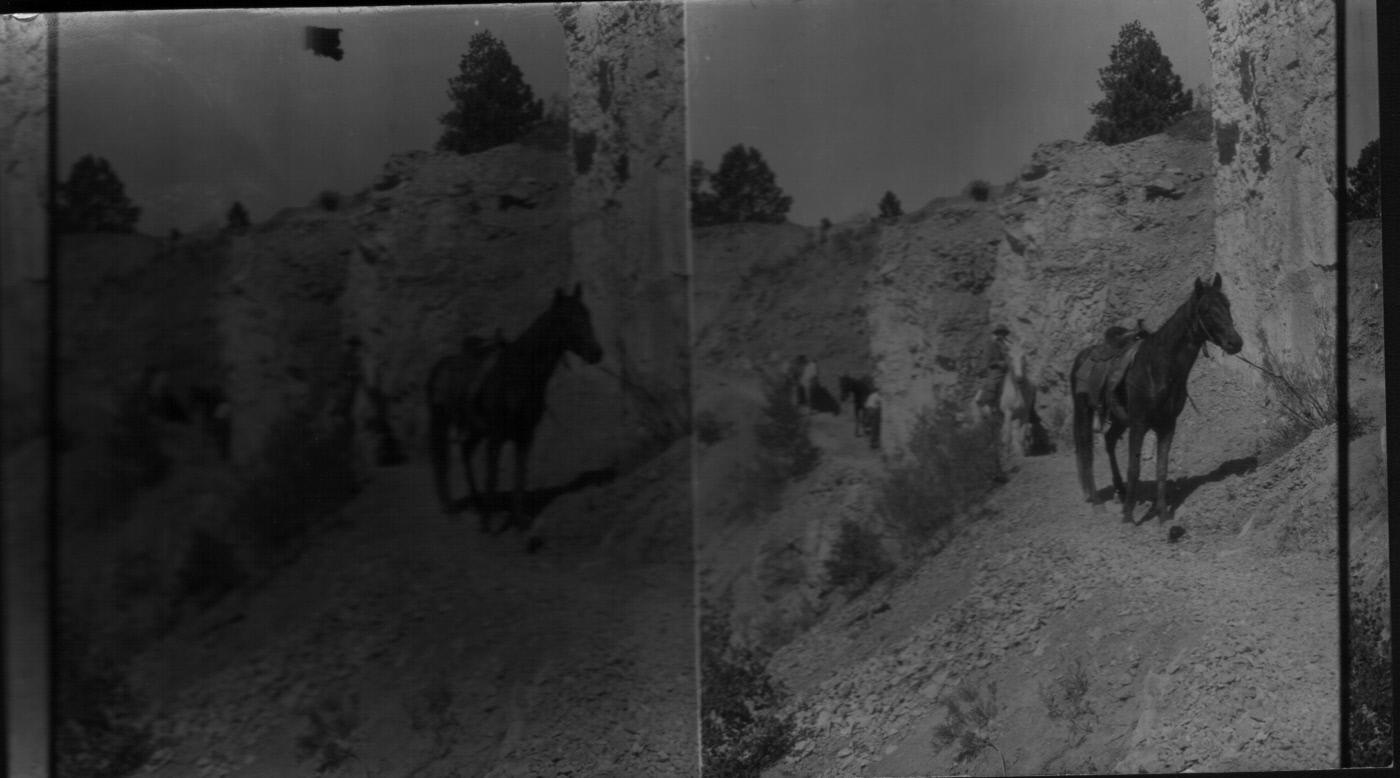} \\
\includegraphics[width=0.48\columnwidth]{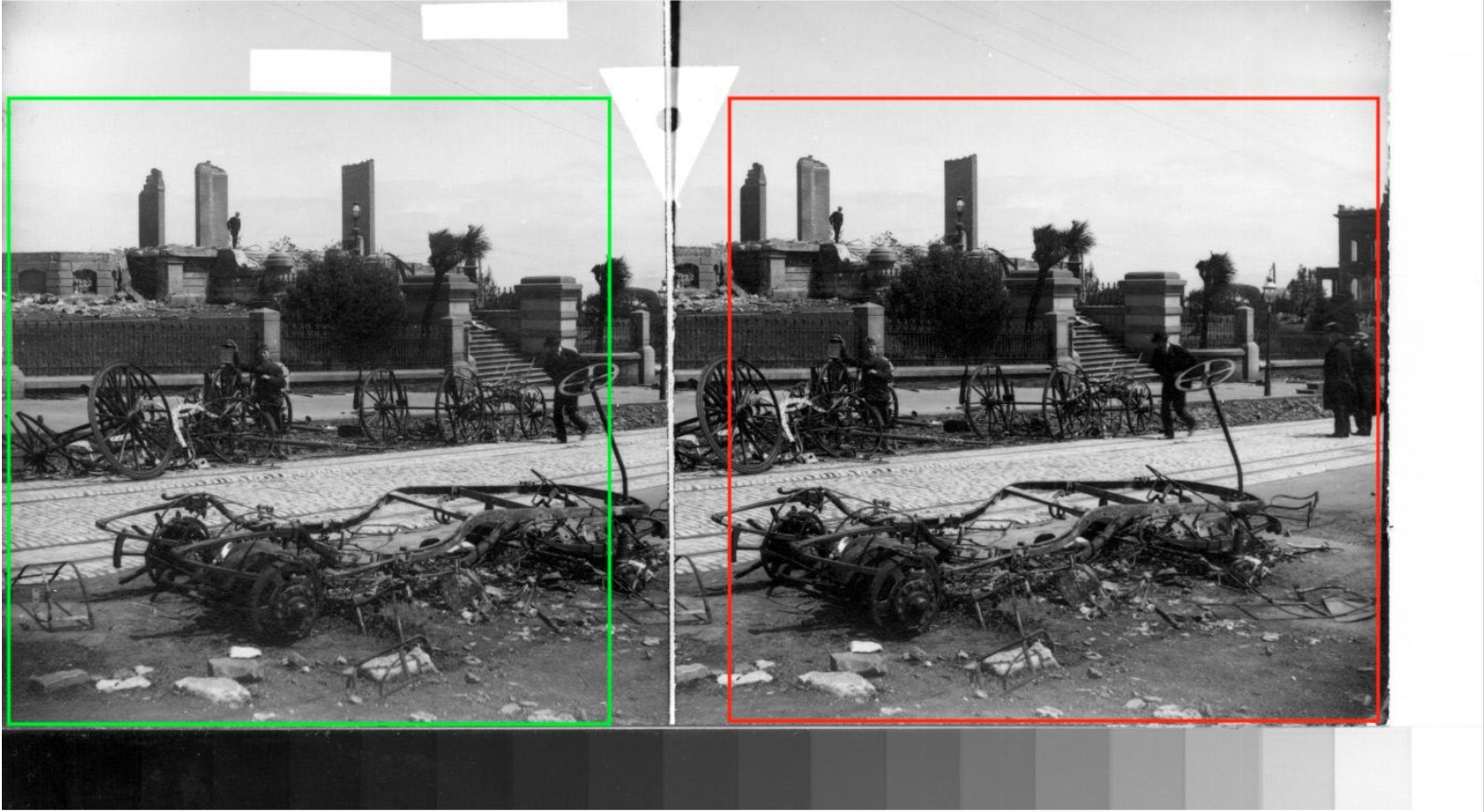} &
\includegraphics[width=0.48\columnwidth]{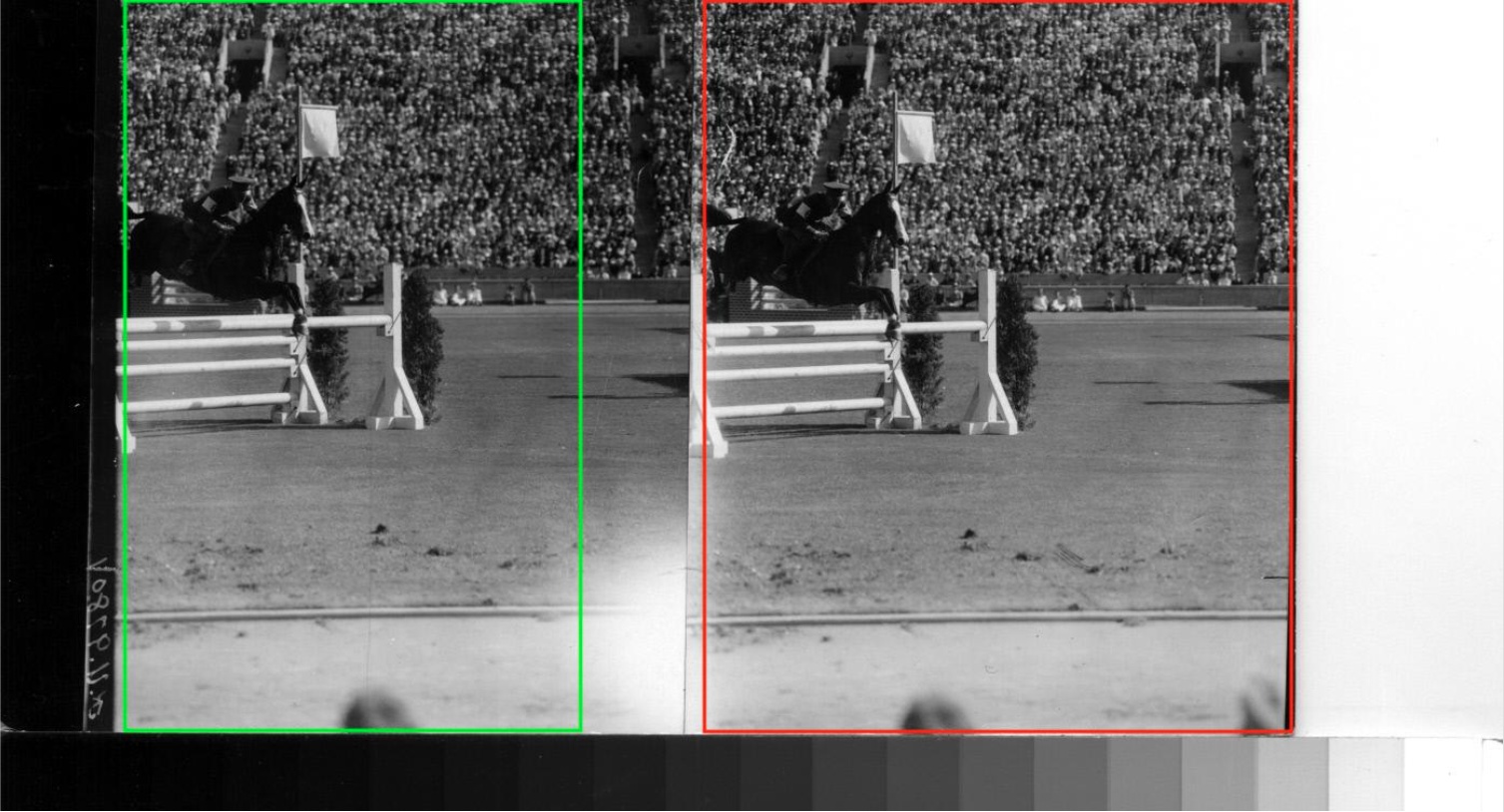} \\
\end{tabular}
\caption{Top row: Two stereo images in the Keystone-Mast collection that were culled due to excessive image artifacts. Bottom row: Two stereo images that were cropped to avoid artifacts. The green and red boxes show cropped areas from the left and right stereo images, respectively.}
\label{fig:bad-images}
\end{figure}

We downloaded stereoscopic images and their accompanying metadata from the Keystone-Mast collection by crawling the index on the CMP website\footnote{\href{http://ucr.emuseum.com/collectionoverview/3631}{http://ucr.emuseum.com/collectionoverview/3631}}.
In a first step, we automatically detected and removed stereo pairs that are unsuitable for depth estimation using the following procedure:
compute SIFT features for both the left and right image and cull any pair having less than $0.5\%$ (or 10 total) good matches between the two sides, where a match is considered "good" if it passes the ratio test with a threshold value of $0.7$.
This step removed most of the monocular images, backs of stereocards, and the poorest quality images. For images with $0.5\% - 1.5\%$ good matches, we hired Mechanical Turkers to inspect each one by hand and identify images that were not valid stereo pairs, upside-down, or had inverted intensities.  We removed the non-stereo images and corrected the latter two cases.

In a second step, we hired a crowdsourcing company~\cite{infosearch} to 
manually specify the largest artifact-free rectangular crop region in each side of a stereo image pair (Figure~\ref{fig:bad-images}, bottom row).
Finally, we used matching SIFT features within these regions to align the crops and compute their intersection. The images were cropped again to this intersection region and processed according to the steps described below.



\begin{figure*}
    \centering
    \captionsetup[subfigure]{aboveskip=1pt,belowskip=0pt}
    \begin{subfigure}[b]{0.205\linewidth}
        \raisebox{10pt}{
            \includegraphics[height=45pt]{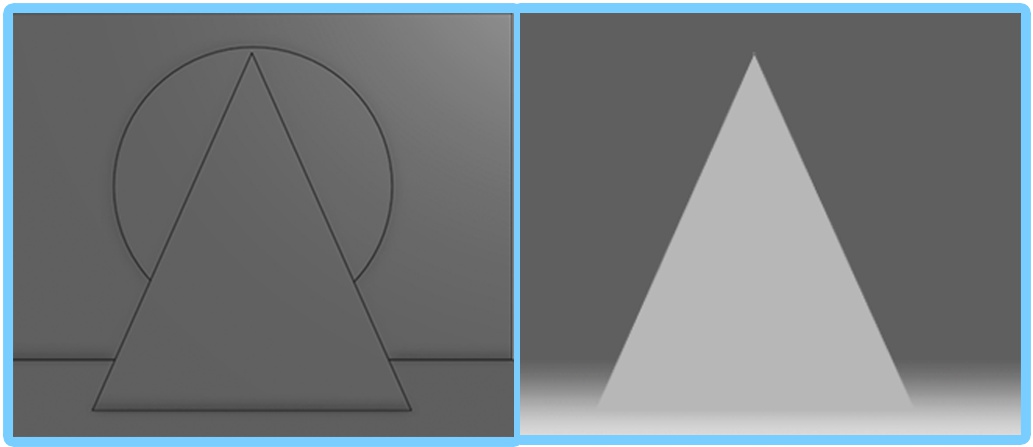}
        }
        \caption{input GD}
    \end{subfigure}\hspace{3pt}%
    \begin{subfigure}[b]{0.245\linewidth}
        \raisebox{10pt}{
            \includegraphics[height=55pt]{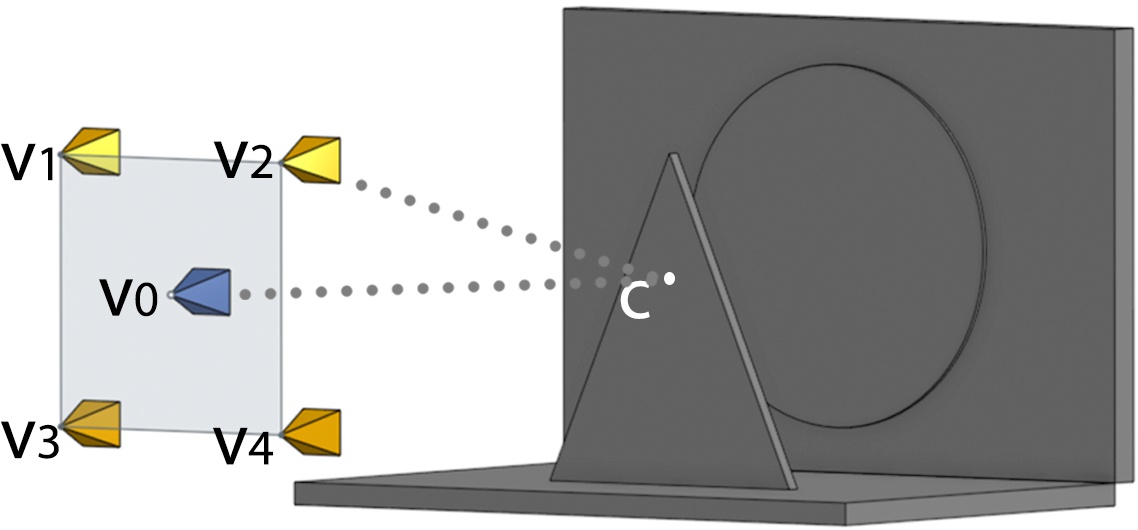}%
        }
        \caption{quad with viewpoints}
    \end{subfigure}\hspace{3pt}%
    \begin{subfigure}[b]{0.265\linewidth}
        \includegraphics[height=70pt]{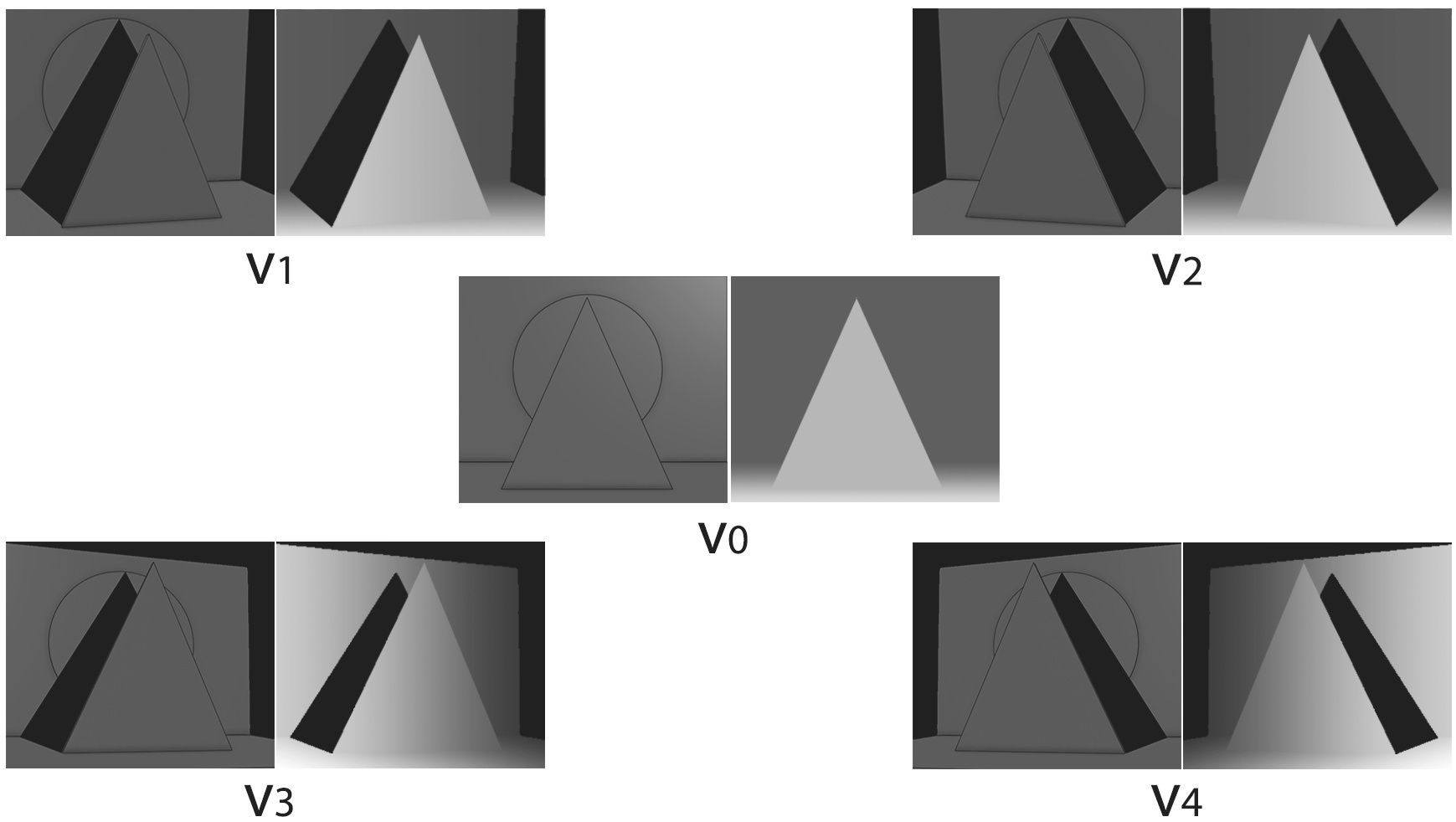}%
        \caption{after reprojection}
    \end{subfigure}\hspace{3pt}%
    \begin{subfigure}[b]{0.265\linewidth}
        \includegraphics[height=70pt]{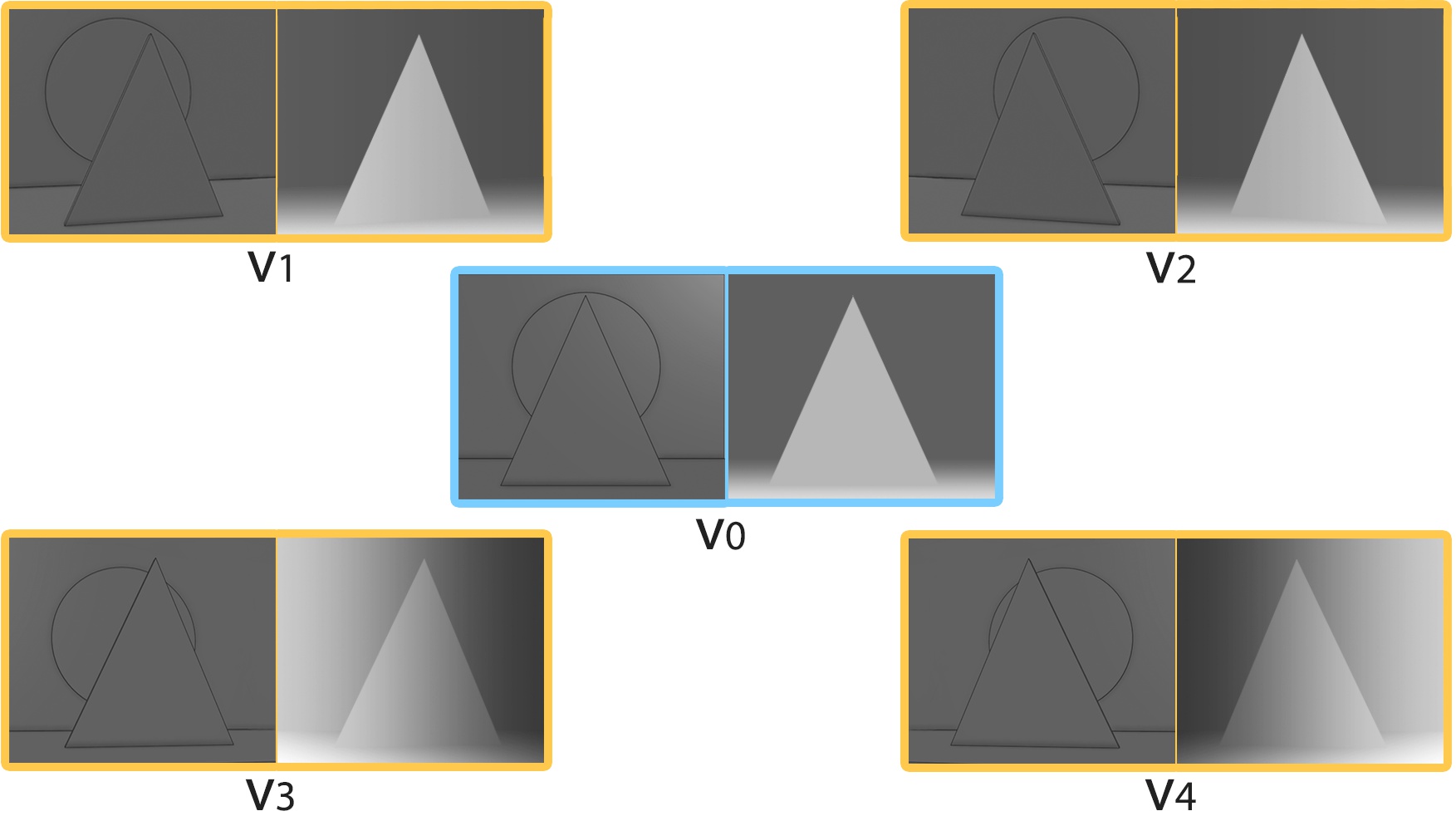}%
        \caption{after inpainting}
    \end{subfigure}%
    \caption{Overview of our scene representation. (a) The input is a single reference grayscale intensity and depth image (``GD image'') computed at the left stereo input image. (b) We compute four new viewpoints at the corners of a quadrilateral surrounding the input image as described in the paper. (c) We reproject the input GD image to these four new viewpoints, which produces holes at object boundaries. (d) We use a self-supervised learning method to inpaint these holes. Our representation of the 3D scene is the union of these GD images, stored as a texture-mapped triangle mesh.}
    \label{fig:representation}
\end{figure*}

\subsection{Stereo Rectification} \label{sec:rectification}

In order to compute disparity maps, each stereo image pair is first rectified so that epipolar lines are aligned with the image pixel rows. 
We used the stereo rectification method developed by Loop and Zhang~\cite{loop1999computing}, which works for uncalibrated cameras, with two modifications. First, the offset of the two principal points in x-axis is ambiguous. We assume the principal point is the center of the full image, if available. Otherwise, we translate the images after rectification so that there is no negative disparity. Second, we assumed a vertical field of view of 45 degrees, as that produced the most plausible depth scaling in our experiments, and inferred a corresponding camera focal length based on this assumption.

\subsection{Disparity Estimation}\label{sec:disp_estimation}

The relatively low quality and noise characteristics of antique stereographs complicates the stereo matching problem, and we found that the top stereo algorithms on the Middlebury benchmark did not produce good results.
Instead, we found that dense optical flow methods performed better, as they are more tolerant of rectification errors that result in vertical displacements between corresponding scene points.

In particular, we used the FlowNet2 algorithm~\cite{ilg2017flownet} to compute optical flow and retained the horizontal component of the flow vector at each pixel as the stereo disparity. We found this strategy produced the best results of the methods we evaluated, producing accurate and sharp boundaries even for inputs as challenging and diverse as those in Figure~\ref{fig:results}.
\section{Real-Time View Synthesis from Antique Stereographs}
\label{sec:overview}

We developed a real-time method for synthesizing novel views from antique stereographs as a way to both visualize the entries in our \emph{KeystoneDepth} dataset and to power a mobile application that gives a user the sensation of looking through a window onto a historical scene.
We first introduce our representation of the 3D scene (Section~\ref{sec:representation}) and then describe a novel learning-based technique for inpainting holes in both depth and intensity images that emerge with viewpoint changes (Section~\ref{sec:inpainting}).

\subsection{Scene Representation}\label{sec:representation}


We seek a compact 3D scene representation that enables rendering new viewpoints within a head volume centered around the input stereo pair at real-time framerates on a mobile device. We first define the supported range of viewpoints as a plane bounded by a quadrilateral (Figure~\ref{fig:representation}).
This quad captures all viewing rays (within the input camera's field of view) that pass through the quad, and thereby enables generating a range of views in front-of and behind the plane, whose rays intersect the quad \cite{levoy96}.

We approximate the continuous range of viewpoints within the quad by sampling five discrete viewpoints: the four corners of the quad, and the center viewpoint (defined by the left image in the stereo pair).  For each of these five viewpoints, we generate a full grayscale intensity plus depth (``GD'') image.  These GD images are treated as textured meshes which are rendered to produce in-between views within the quad.  Unlike most traditional view interpolation methods, which interpolate new views from real photos,
these corner GD images are {\em synthesized}.  As such we perform two phases of view synthesis:  one to extrapolate these quad corner views (offline), and a second to interpolate views inside this quad (online in real-time).

Formally, take the left image of the stereo pair as a reference, and assume depth has been computed to form a reference GD image denoted $v_0$.
Consider a coordinate system whose origin is located at the reference camera's center of projection at $v_0=(0,0,0)$ with its -z direction aligned with the optical axis of the reference camera, as shown in Figure~\ref{fig:representation}. We pick the scene center to be $c=(0,0,-1/median(1/D_0))$, where $D_0$ is the reference depth map. Based on the head movement $(r_w, r_h)$ we want to support, we set the quad corners to $v_1=(-r_w,r_h,0)$, $v_2=(r_w, r_h, 0)$, $v_3=(-r_w, -r_h, 0)$, and $v_4=(r_w, -r_h, 0)$. Further, we orient the cameras at the corners towards the scene center. Let $R(v,c)$ denote the rotation matrix for a camera located at $v$ looking at the location $c$ with up vector the same as the reference camera. Our scene representation thus consists of the set of image tuples $\langle I(v_i, R(v_i, c)), D(v_i, R(v_i,c))\rangle$, $i\in[0,4]$, where $I(v,R)$ and $D(v,R)$ are intensity and depth images from a camera at position $v$ and rotation $R$, respectively. 


Choosing appropriate values for $r_w$ and $r_h$ is application dependent, as it depends on the depth of the scene contents and the desirable range of head motion. In the case of antique stereographs, however, we generally don't have access to the metric camera parameters, and therefore cannot specify physically meaningful viewpoint ranges.  Therefore, we instead choose these parameters based on the desired amount of image parallax when moving from one side of the quad to the other.
Specifically, we obtained good results on average with $r_w=r_h=\frac{96b}{d_{max}}\times\frac{\sqrt{2}}{2}$, where the $d_{max}$ is the maximum disparity and $b$ is the camera baseline, corresponding to maximum displacement of 96 pixels.
\section{Intensity and Depth Inpainting}\label{sec:inpainting}

Our scene representation requires reconstructing intensity and depth images of the 3D scene as seen from the four quad corners. For most scenes, reprojecting the reference GD image (left stereo input image) to these viewpoints will produce images that have holes along depth boundaries where parts of the scene come into view that are not visible from the reference viewpoint (Figure~\ref{fig:representation}). A fundamental problem that any view synthesis method must overcome is filling in this type of missing data.

We take a machine learning approach to this problem and train a neural network to inpaint missing regions in GD images of these types of scenes, and build on state-of-the-art inpainting techniques~\cite{partial-convolution} designed for color images. 
The method of Liu et al.~\cite{partial-convolution} uses training data augmented by adding holes, in the form of rectangles or random streaks. We observe that holes in reprojected images are different and have a characteristic structure: they follow object {\em boundaries} and correspond primarily to {\em background} regions in the scene that become visible due to disocclusions.
One way to produce training data that possesses this structure is to reproject a GD image $v$ to a new viewpoint $v'$ and compare it to a ground truth image from $v'$.  
However, such an approach requires having ground truth depth and intensity images, which is not available in our case.
We therefore introduce a technique that generates realistic holes across a single complete GD image and apply it to train a neural network for this inpainting task.

\subsection{Double Reprojection}\label{sec:self-supervision}

\begin{figure}[t]
\centering
\setlength{\tabcolsep}{0.35mm}
\def\imw{0.32\linewidth}
\begin{tabular}{ccc}
        \includegraphics[width=\imw]{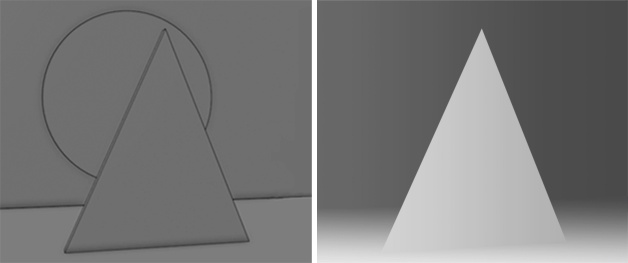}
    &
        \includegraphics[width=\imw]{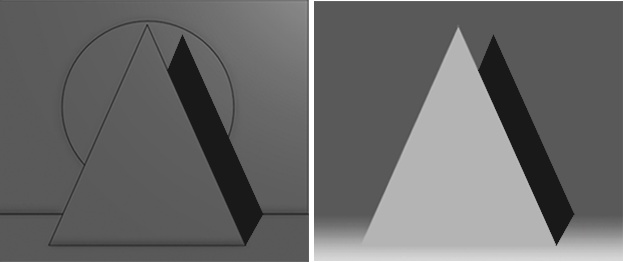}
&
        \includegraphics[width=\imw]{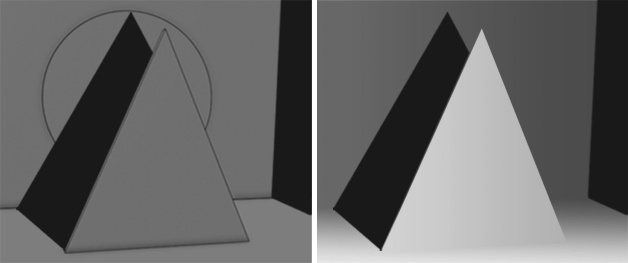}
\\ $v$ & $v'$ & $v$
\end{tabular}
\vspace*{4pt}
    \caption{We reproject a GD image at viewpoint $v$ to a different viewpoint $v'$ and then back to $v$, in order to generate a pattern of holes at $v$ that are characteristic of those revealed by viewpoint changes. We apply this {\em double reprojection} to train a neural network to inpaint these types of holes.}
    \label{fig:double-reprojection}
\end{figure}

As shown in Figure~\ref{fig:double-reprojection}, reprojecting an input GD image from $v$ to a new viewpoint $v'$ and then back to $v$ produces an image that is comparable to projecting a complete GD at $v'$ to $v$. However, the advantage of the former is that it requires only one GD, not two.
Hence, we can produce a large number of training images with holes and hole-free ground truth by performing this {\em double reprojection} using the set of GD images in the {\em KeystoneDepth} dataset.

Specifically, we construct a triangle mesh from the input depth image that is textured using the image intensities and remove triangles that straddle depth boundaries by thresholding maximum relative difference in depth. We then render the resulting mesh from each of the viewpoints at the quad corners. Finally, we reconstruct a new GD image from each rendered result (now having holes) and perform this meshing-and-rendering operation a second time, but now from the original viewpoint. The result is a set of masks that mark characteristic hole locations in the original GD image.

\subsection{Boundary mask}
We found that inpainting intensity and depth across depth boundaries tends to generate blurry transitions between the foreground and background parts of the scene. 
To avoid these kind of artifacts, we add a boundary mask as an additional input to the network, that indicates the location of depth discontinuities.
In general, one can only infer a depth discontinuity on the foreground side of a reprojection hole, as the scene might have constant depth along the background side of it.
Therefore, we detect pixels on the foreground side of the reprojection holes and store them as a binary boundary mask $B$.
Sample boundary masks are shown in Figure~\ref{fig:boundary_guidance}, and we provide more details on how they are computed in the supplementary material.

\subsection{Inpainting network}
We train a deep convolutional neural network to inpaint holes in GD images. The architecture of our network is the same as that proposed by Liu et al~\cite{liu2018image} for similar image synthesis tasks: a standard U-Net~\cite{ronneberger2015u} where the convolutional layers have been replaced with partial convolutions.  

\paragraph{Network Inputs and Outputs:}
Each sample in our training set is given by $\langle I, \hat{D}, B, M\rangle$, where $I$ is the intensity image, $\hat{D}$ is the normalized inverse depth, $B$ is the boundary mask, and $M$ is a binary mask corresponding to the holes that are to be inpainted. We define the normalized inverse depth as $\hat{D}=[(1/D)-D_{m}]/(D_{M}-D_{m})$, where $D_{m}=\min(1/D), D_{M}=\max(1/D)$. The inputs to our network are then $( I\odot M, \hat{D}\odot M, B\odot M),$ where $\odot$ denotes element-wise multiplication. We trained two networks, one for inpainting intensity holes and one for depth, whose outputs are $\hat{I}$ and $\hat{D}$ respectively.

\paragraph{Losses:}
For inpainting image intensities, we used the same loss as Liu et al.~\cite{partial-convolution} including $L1$ loss in both valid and hole regions, total variation loss, perceptual loss and style loss.
For depth inpainting, we used an $L1$ loss with different weigths for the valid (non-hole) and hole regions, and a total variation loss on the composite depth, to encourage sparse depth discontinuities in the final output, i.e.,
\[
    L(\hat{D}_p,\hat{D}_t;M) = L_{valid}+\lambda_{hole}L_{hole}+\lambda_{tv}TV(\hat{D}_{comp}\odot M)
\]
\[
    L_{valid} = ||(\hat{D}_p-\hat{D}_t)\odot M||_1 \]
\[
    L_{hole} = ||(\hat{D}_p-\hat{D}_t)\odot(1-M)||_1
\]
where $\hat{D}_p$ and $\hat{D}_t$ are predicted and ground truth normalized inverse depth and $TV(x)$ is the total variation loss, and $\hat{D}_{comp} = (1-M)\odot \hat{D}_p + M\odot \hat{D}_t$ is their composition. 


\subsection{Implementation Details}

We train the intensity and depth inpainting network on a subset of our dataset containing 676 samples from the \emph{KeystoneDepth}. To improve results, we pre-train on a dataset of 9400 samples of SUNCG~\cite{zhang2016physically}, where we create characteristic holes using the double reprojection technique.

We trained the network using the ADAM solver~\cite{adam}. We pre-train for 125K steps on SUNCG, and then train 17K steps on KeystoneDepth data. In both cases, we set $\beta_1 = 0.9$, $\beta_2 = 0.999$ for ADAM solver, and  use a batch size if $4$. The learning rate is set to $2\times 10^{-3}$ during pre-training and $2\times 10^{-4}$ when training on KeystoneDepth. During training, all images have a spatial resolution of $512 \times 512$. However, because our network is fully convolutional, it can be applied to images of arbitrary resolution. Training required about two days with a Nvidia Tesla 1080Ti GPU. 

We use weights $\lambda_{hole}=6,\lambda_{tv}=0.1$, and for the double reprojection technique, we discard a triangle if two of its vertices have a relative depth difference $> 0.1$.
\section{Evaluation}

We performed a number of qualitative and quantitative evaluations of our \emph{KeystoneDepth} dataset and novel view synthesis technique. Throughout our experiments, we set $r_w=r_h=\frac{96b}{d_{max}}\times\frac{\sqrt{2}}{2}$ for inpainting, and generate novel views with a head volume of $[-r_w/4,r_w/4]\times[-r_h/4, r_h/4]\times[-1.5r_w,0]$.

\begin{figure}
    \centering
    \includegraphics[height=0.4\linewidth]{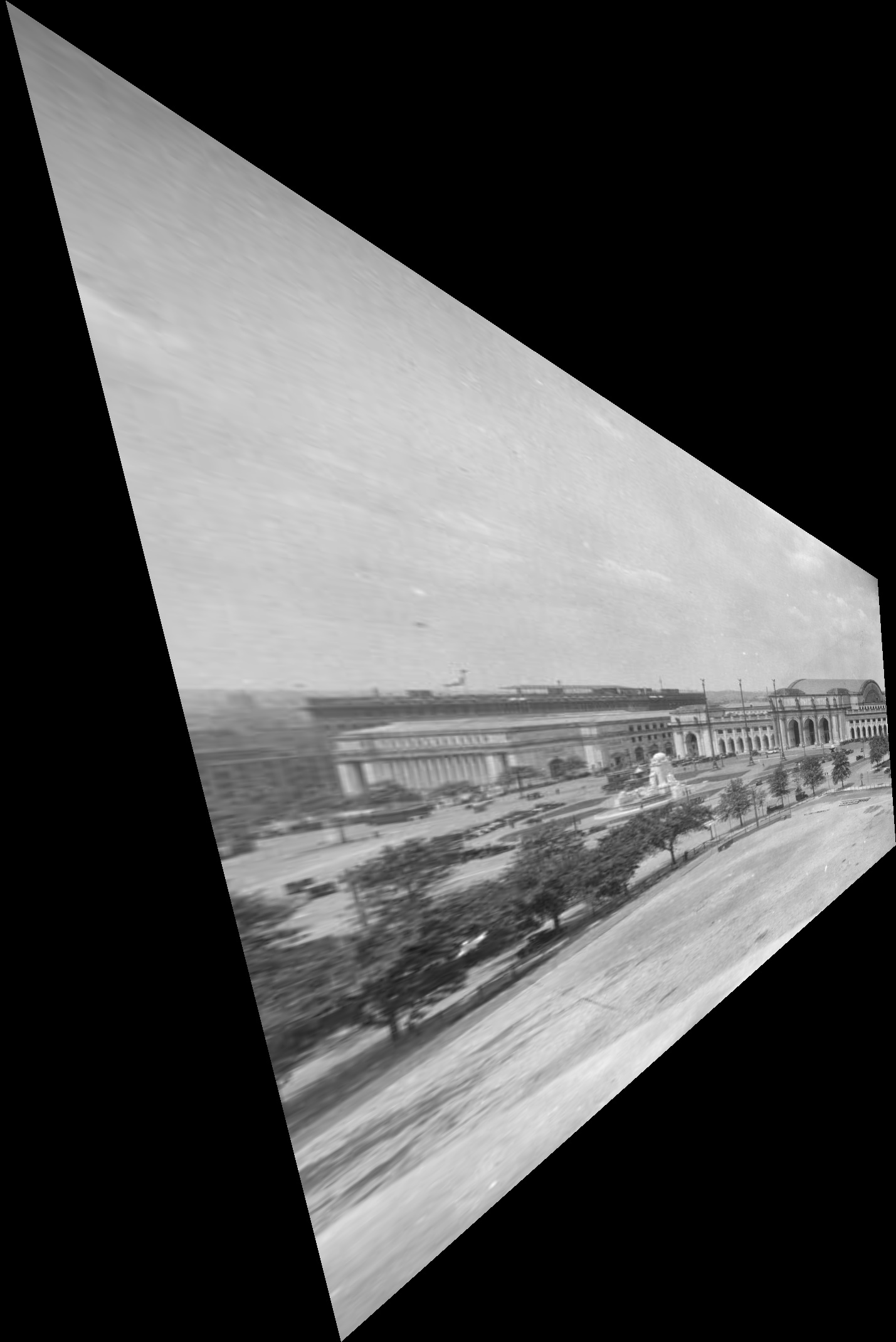}\hspace{2pt}%
    \includegraphics[height=0.4\linewidth]{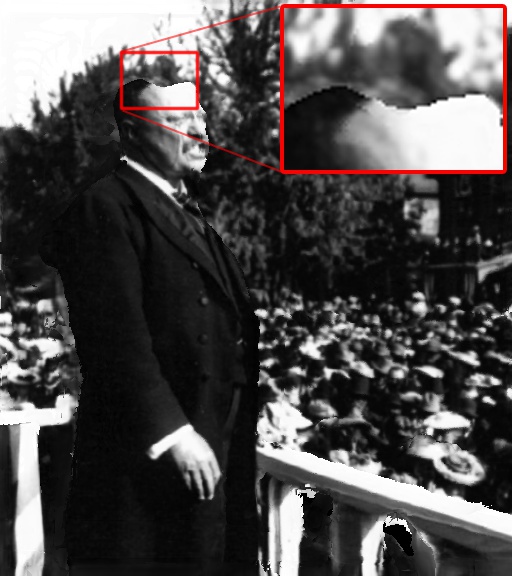}\hspace{2pt}%
    \includegraphics[height=0.4\linewidth]{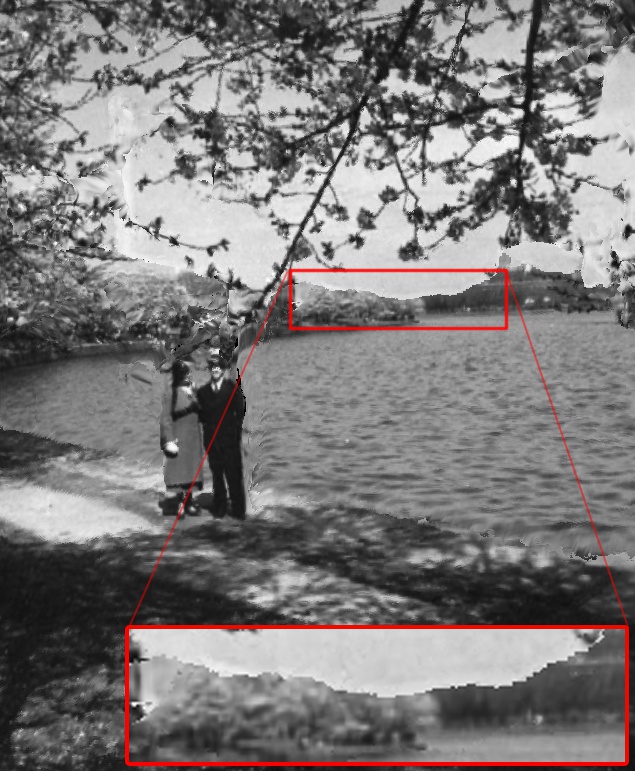}
    \caption{Example failure cases. Left: Rectification failed due to negligible disparity of distant objects and lens distortion. Middle: Errors in optical flow in person's hair because its color is similar to the trees behind. Right: Depth estimate of the sky is incorrect (too close to viewer) due to limited texture and poor optical flow.}
    \label{fig:failure_cases}
\end{figure}

\begin{figure*}
\centering
\renewcommand{\arraystretch}{1}
\setlength{\tabcolsep}{2pt}
\def\imw{0.211\linewidth}
\begin{tabular}{cccc}
    \includegraphics[trim={0 0.5cm 0 1cm},clip,width=\imw]{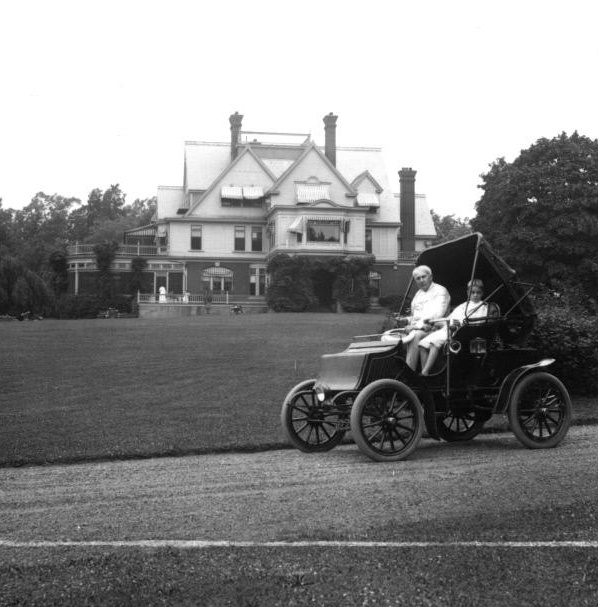} & \includegraphics[trim={0 0.5cm 0 1cm},clip,width=\imw]{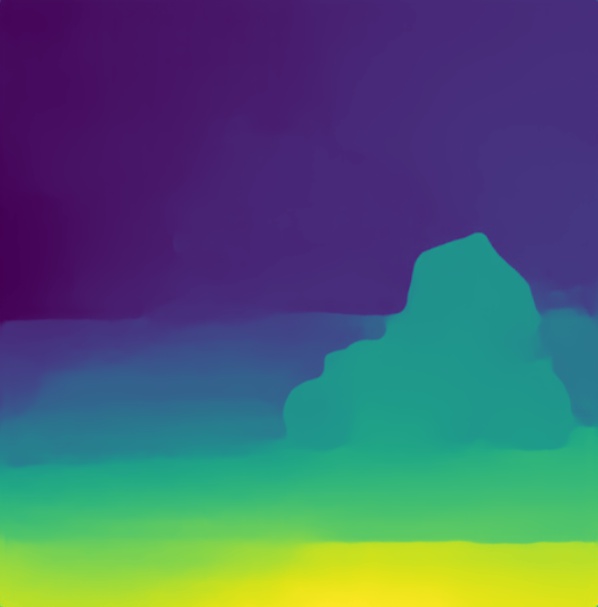} & \includegraphics[trim={0 0.5cm 0 1cm},clip,width=\imw]{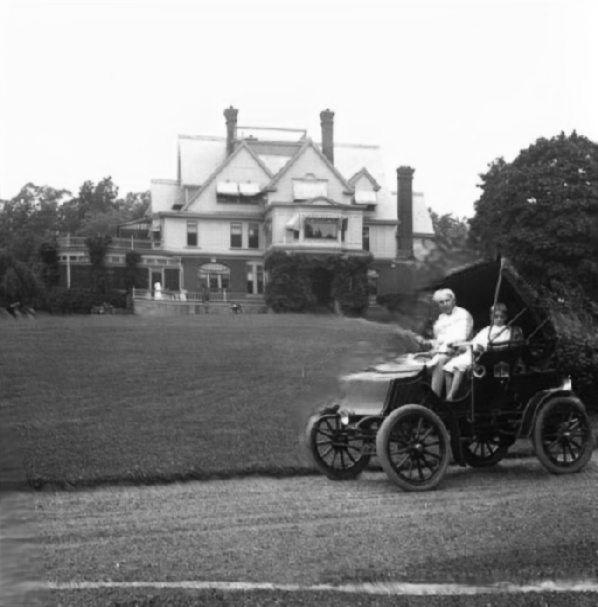} & \includegraphics[trim={0 0.5cm 0 1cm},clip,width=\imw]{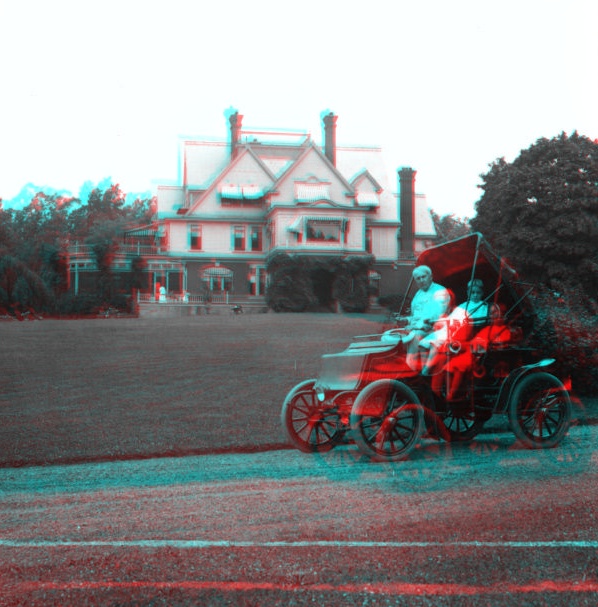} \\
    \includegraphics[width=\imw]{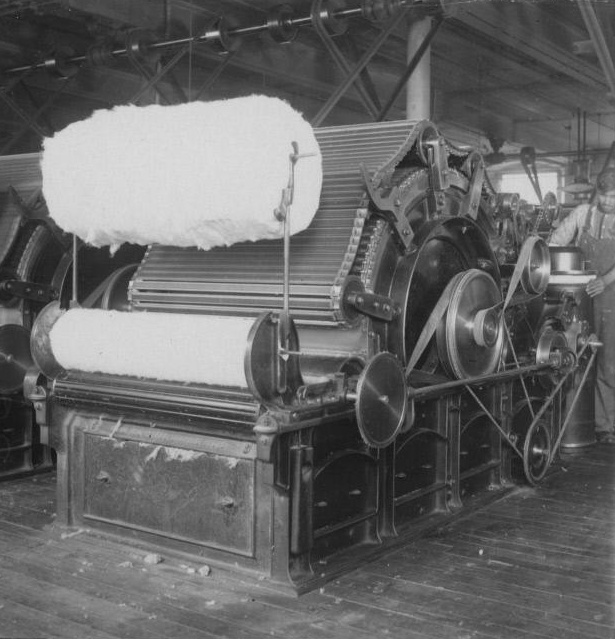} & \includegraphics[width=\imw]{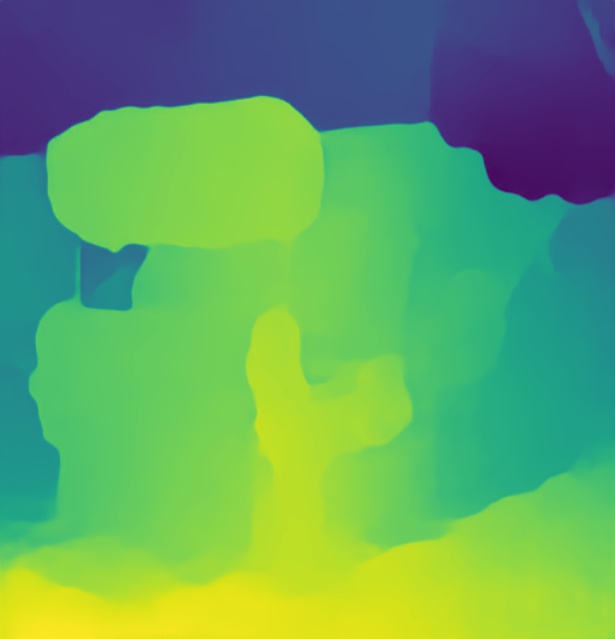}  & \includegraphics[width=\imw]{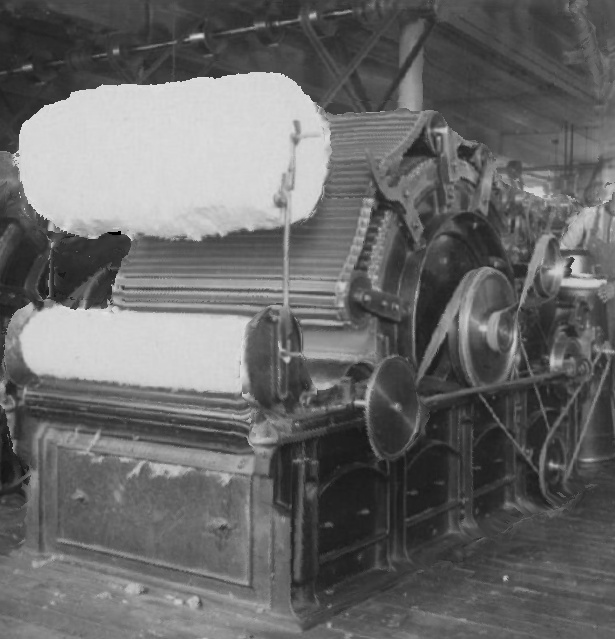} & \includegraphics[width=\imw]{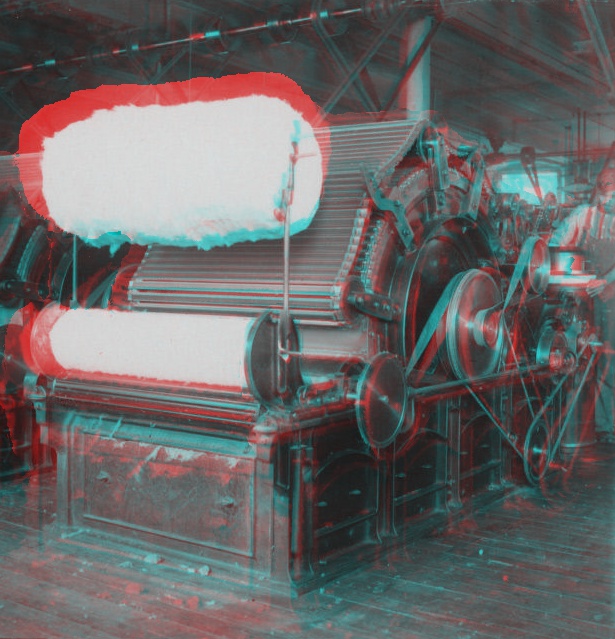} \vspace*{1pt} \\
    \includegraphics[trim={0 0.5cm 0 2cm},clip,width=\imw]{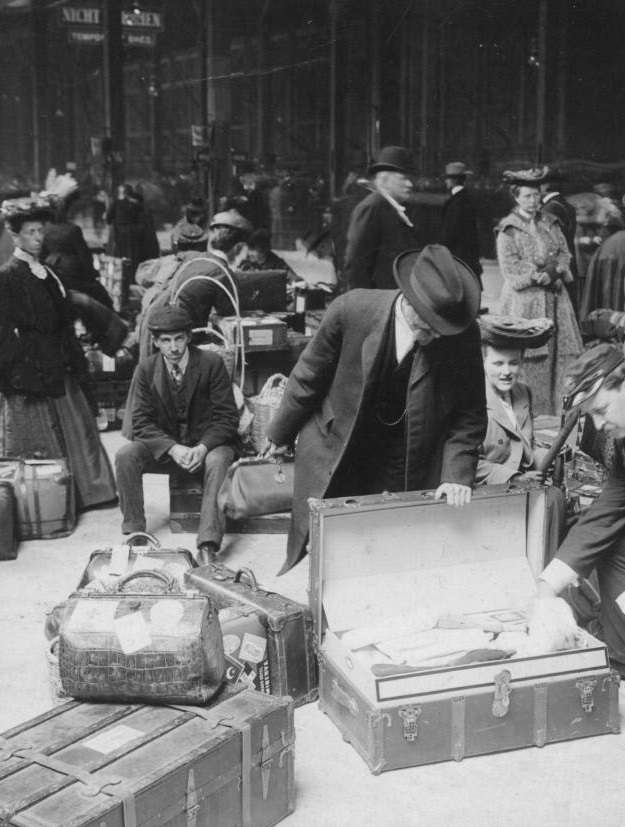} & \includegraphics[trim={0 0.5cm  0 2cm},clip,width=\imw]{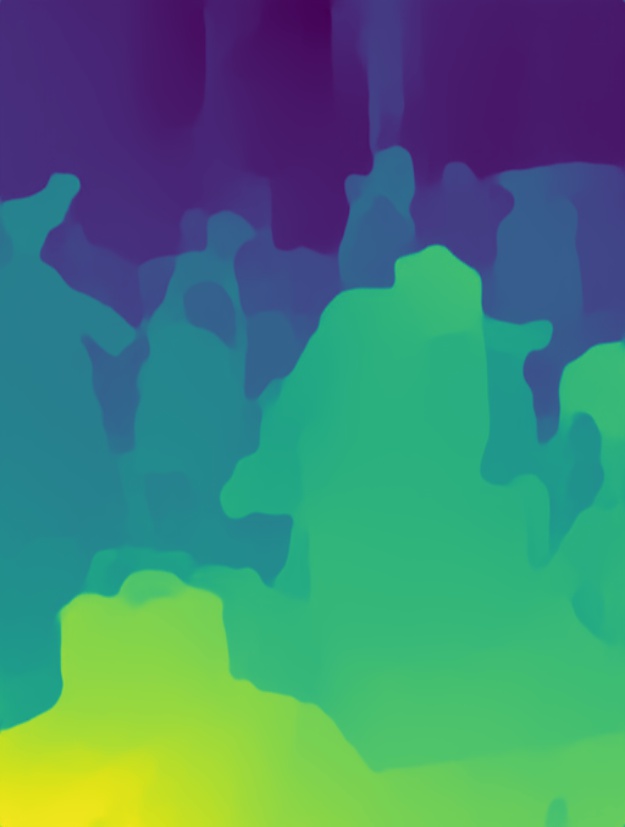}  & \includegraphics[trim={0 0.5cm 0 2cm},clip,width=\imw]{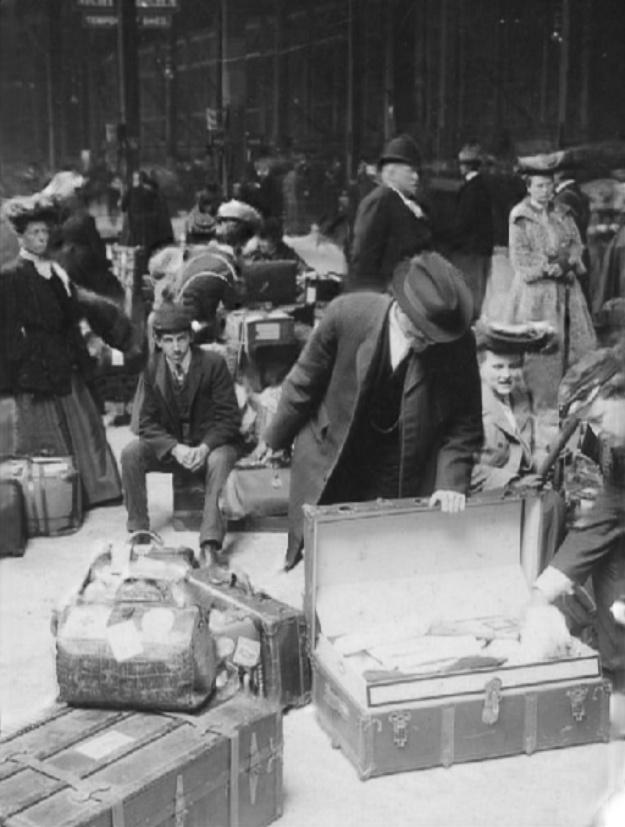} & \includegraphics[trim={0 0.5cm  0 2cm},clip,width=\imw]{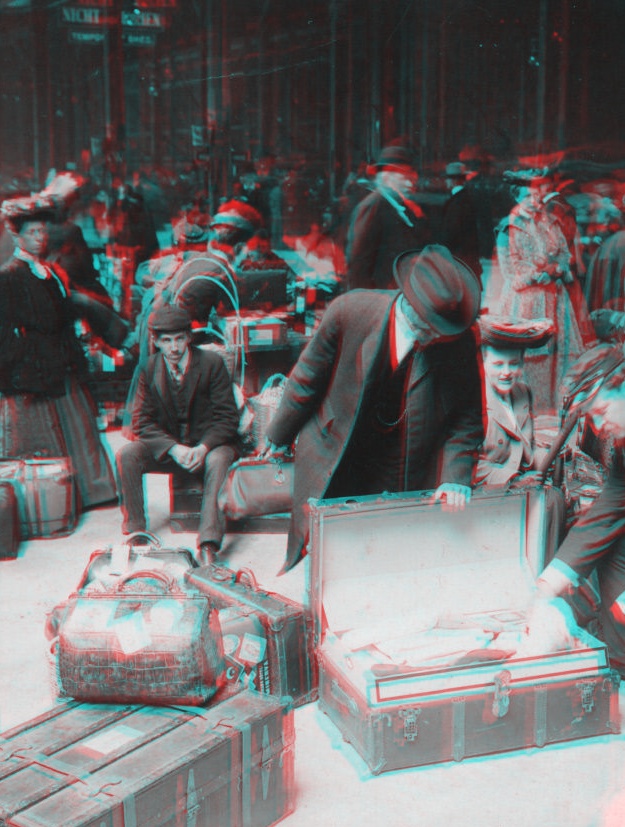} \\
    \includegraphics[width=\imw]{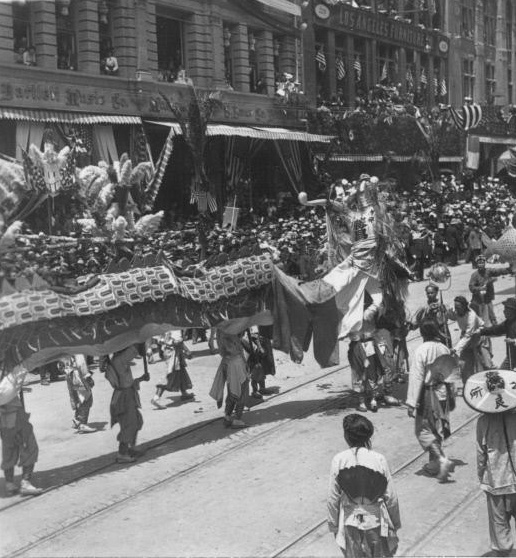} & \includegraphics[width=\imw]{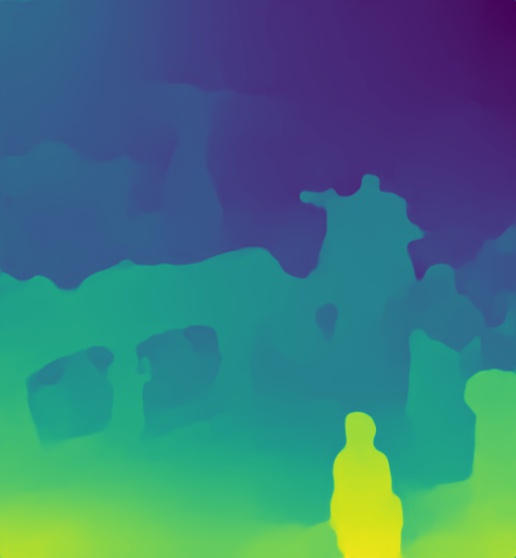}  & \includegraphics[width=\imw]{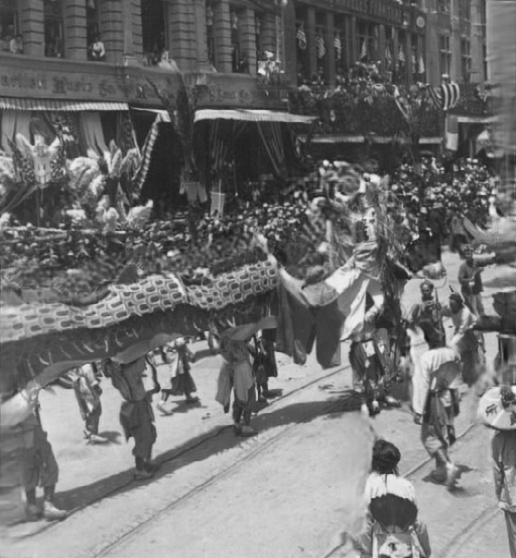} & \includegraphics[width=\imw]{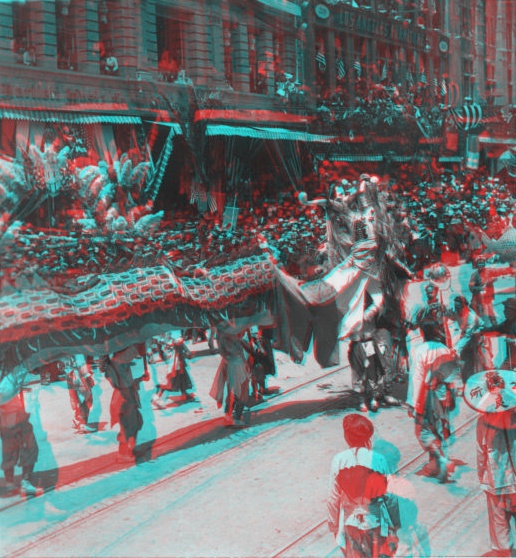}  \\
    \includegraphics[width=\imw]{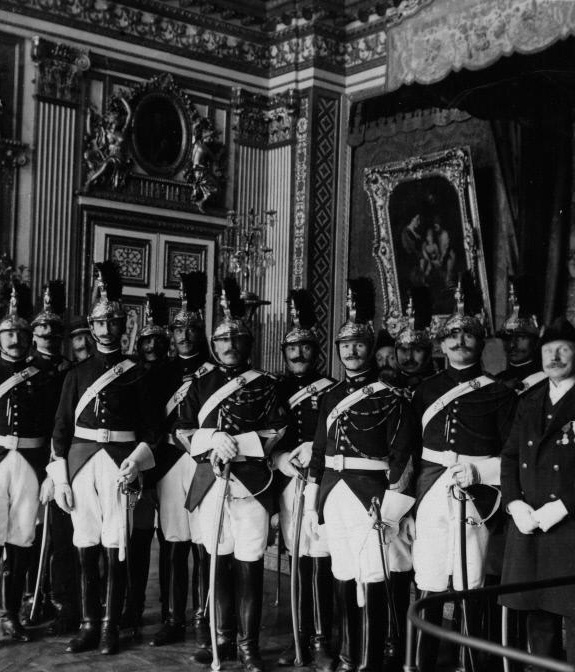} & \includegraphics[width=\imw]{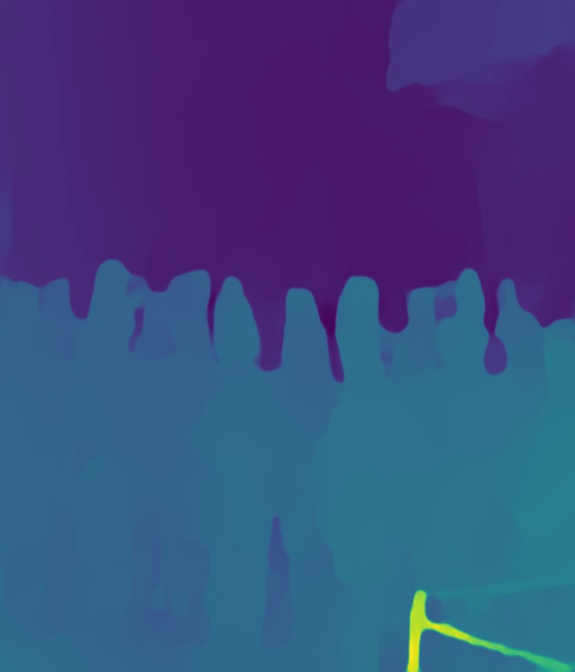}  & \includegraphics[width=\imw]{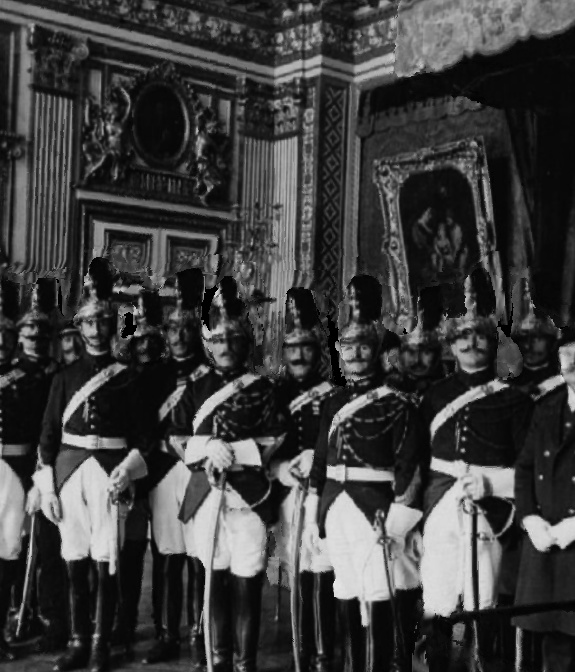} & \includegraphics[width=\imw]{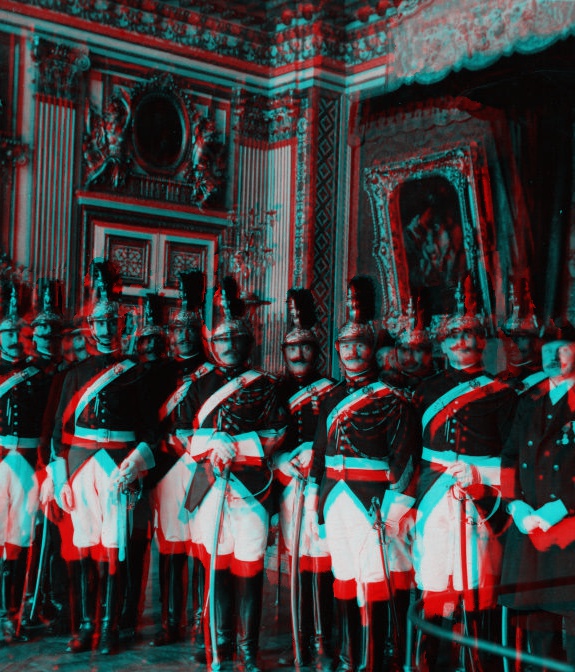} \vspace*{-1pt} \\
    (a) & (b) & (c) & (d)
\end{tabular}
\caption{(a) Left stereo input image. (b) Depth map. (c) Reconstructed view at the following positions (top to bottom): $v_1$, a top-left corner in front of $v_0$, $v_2$, $v_1$, a top-right corner in front of $v_0$. (d) Anaglyph showing the composite of (a) and (c) to help illustrate the parallax in the scene. From top to bottom: \emph{Thomas Edison}, \emph{Cotton Carding Machine}, \emph{Examining Baggage at Ellis Island}, \emph{Chinese Celestial Dragon}, and \emph{Guards at Signing of Treaty of Versailles}. Please see supplementary material for video versions of these results and many more.}
    \label{fig:results}
\end{figure*}

\paragraph{Dataset:} Figure~\ref{fig:results} shows a few of the more than 10,000 stereo images in the \emph{KeystoneDepth} dataset.
Please see the supplementary material which includes video versions of these and hundreds of additional results.

To evaluate the quality of our results, we manually categorized 711 randomly sampled entries from our dataset and assigned a label to each: "very few artifacts", "some artifacts", and "failure." We found $23\%$ to have very few artifacts, $49\%$ show some artifacts, and the remaining $28\%$ are failures. Extrapolating these statistics to the entire collection implies there are several thousand high quality results. Figure~\ref{fig:failure_cases} shows representative failure cases due to poor rectification and poor depth estimation. The supplementary material also includes additional examples in each of these quality categories.

Among the failure cases, we observe that about $5\%$ are caused by incorrect rectification or because the stereo cards have the eyes swapped.
Improving the processing to achieve even higher quality results over a larger part of the Keystone-Mast collection is an interesting direction for future work and one we believe will become possible by making this dataset available to the research community.

\paragraph{Inpainting network:}
\begin{figure}[ht]
    \centering
    \includegraphics[width=0.32\linewidth]{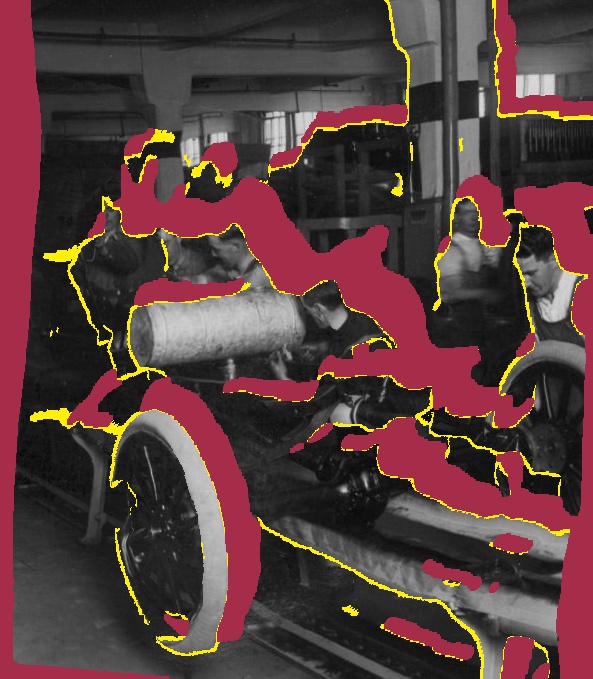} \includegraphics[width=0.32\linewidth]{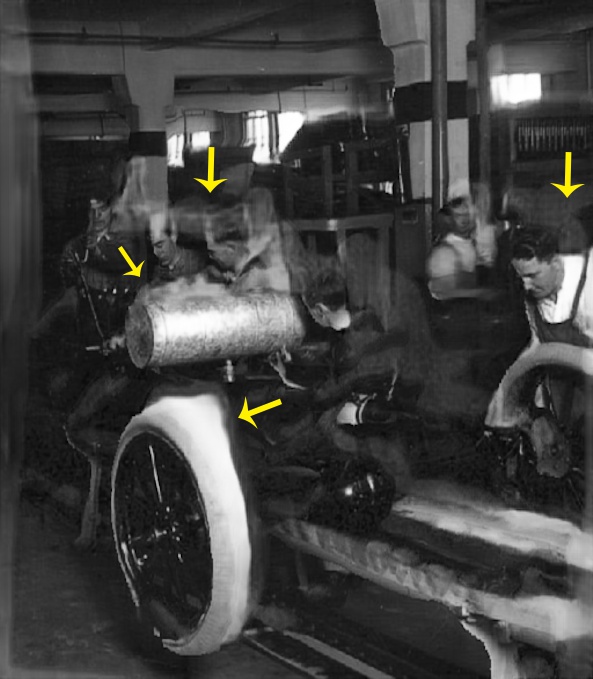} \includegraphics[width=0.32\linewidth]{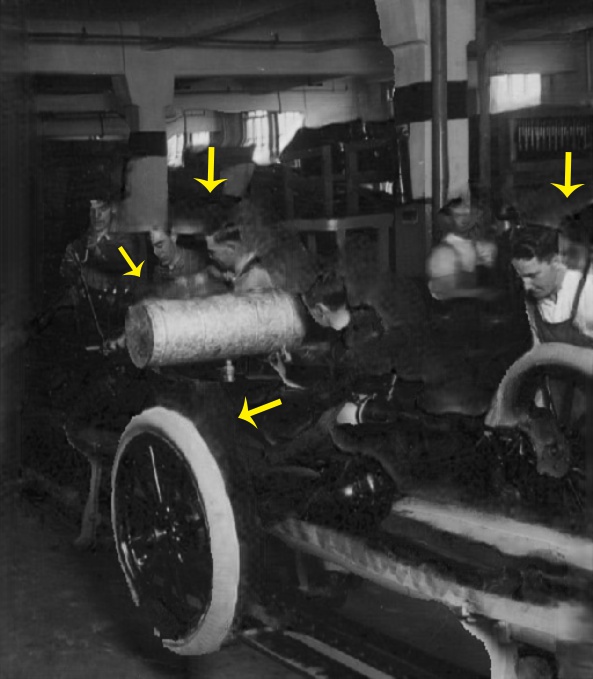}
    \caption{Left: Object boundary mask (yellow) superimposed on an intensity image with holes marked in red. Middle: Inpainting result without boundary guidance. Right: Result with boundary guidance.}
    \label{fig:boundary_guidance}
\end{figure}

\begin{figure}[b!]
    \centering
    \includegraphics[width=0.32\linewidth]{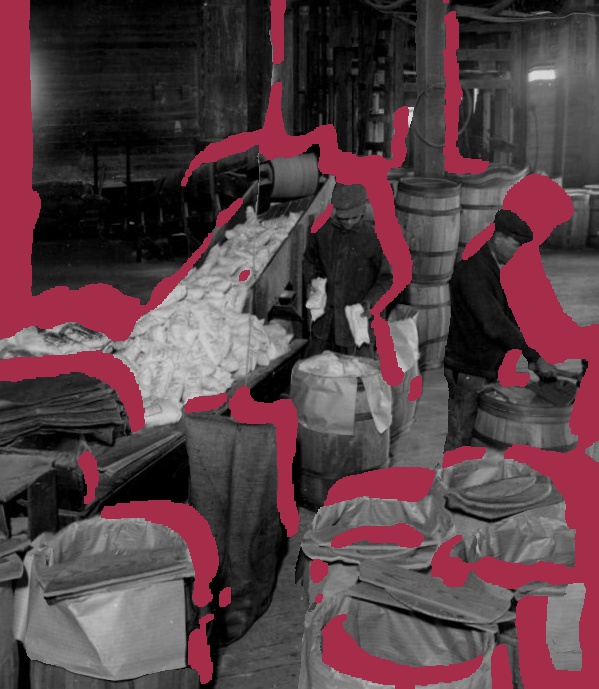} \includegraphics[width=0.32\linewidth]{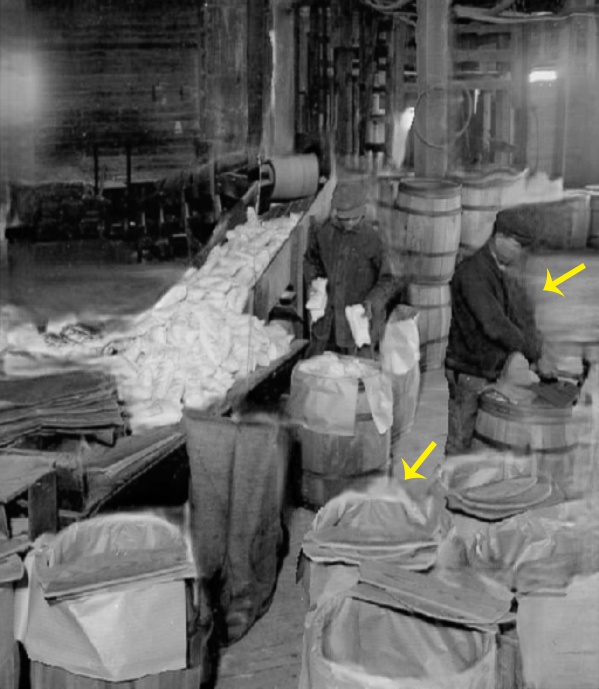}  \includegraphics[width=0.32\linewidth]{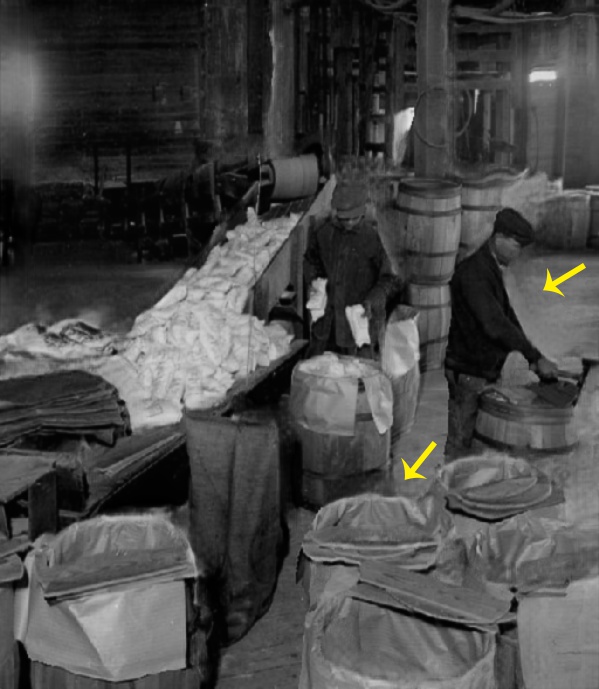}
    \caption{Results of training with and without double reprojection. Left: Input holes in dark red. Middle: Results without double reprojection. Right: Results with double reprojection. }
    \label{fig:double-reprojection-comparison}
\end{figure}

We also evaluated how the unique components of our neural network for inpainting contribute to its overall performance. Figures~\ref{fig:boundary_guidance} and~\ref{fig:double-reprojection-comparison} show the impact of including the boundary mask as an additional input and using double reprojection technique for training, which guide the network to produce sharper and more continuous depth boundaries.

\begin{figure}
    \centering
    \renewcommand{\arraystretch}{1}
    \newcolumntype{P}[1]{>{\centering\arraybackslash}p{#1}}

    \setlength{\tabcolsep}{1pt}
    \begin{tabular}{P{0.39\linewidth}P{0.39\linewidth}P{0.19\linewidth}}
    \multirow{2}{*}[0.65in]{\includegraphics[width=\linewidth]{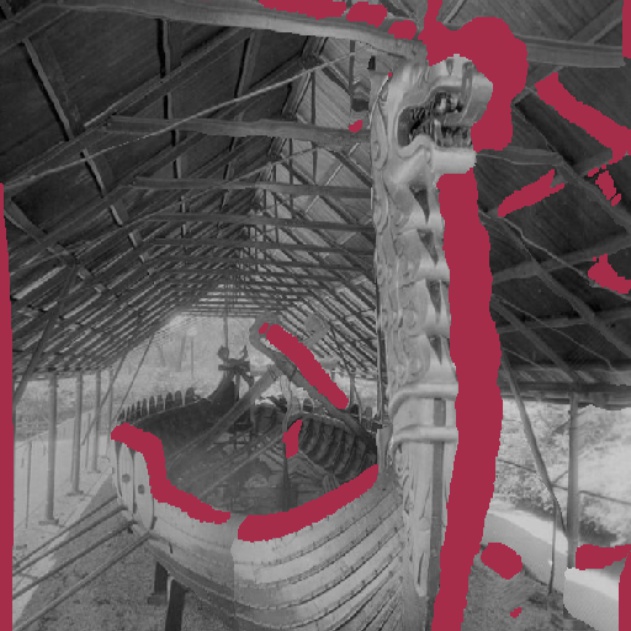}} &  \multirow{2}{*}[0.65in]{\includegraphics[width=\linewidth]{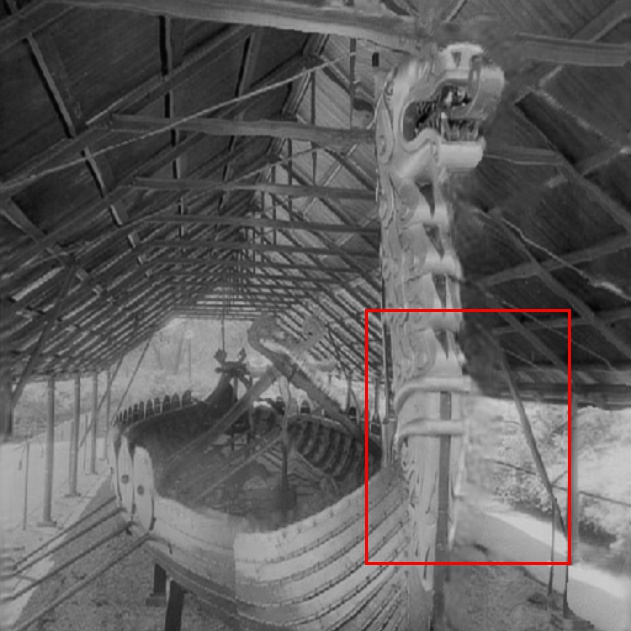}} &
    \hspace{-12pt}\raisebox{11.3pt}{
    \includegraphics[width=0.8\linewidth]{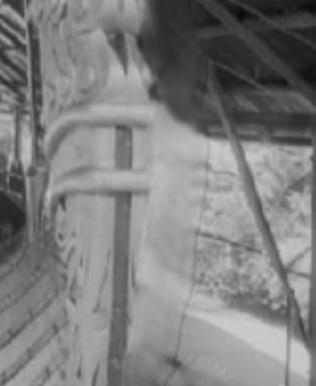}} \\
    & &\vspace{-18.8pt}\hspace{-12pt}
    \includegraphics[width=0.8\linewidth]{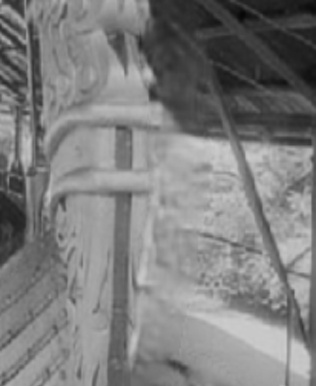} \\
    \multirow{2}{*}[0.65in]{\includegraphics[width=\linewidth]{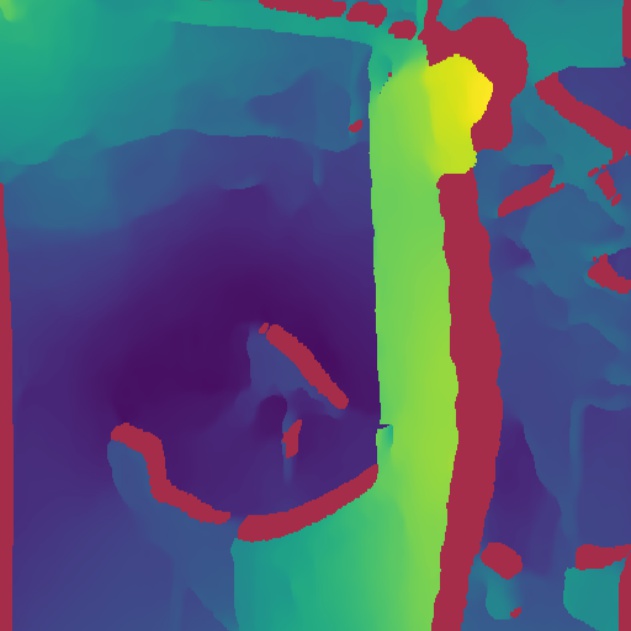}} & 
    \multirow{2}{*}[0.65in]{\includegraphics[width=\linewidth]{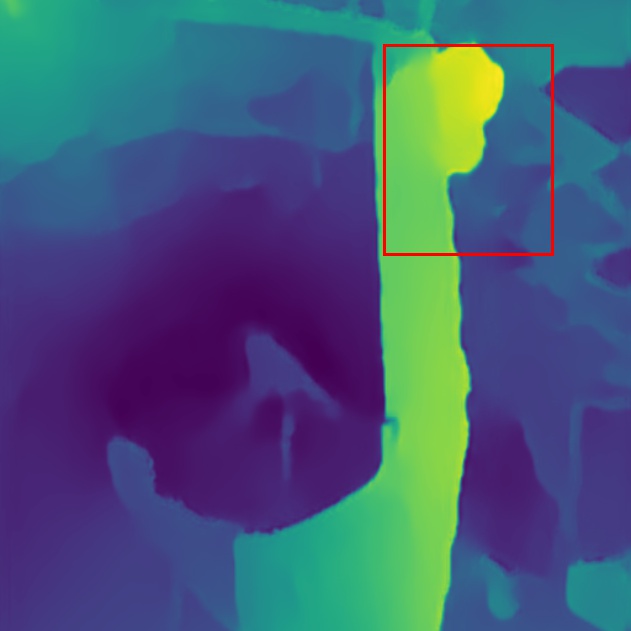}} & \hspace{-10pt}\raisebox{11.3pt}{\includegraphics[width=0.8\linewidth]{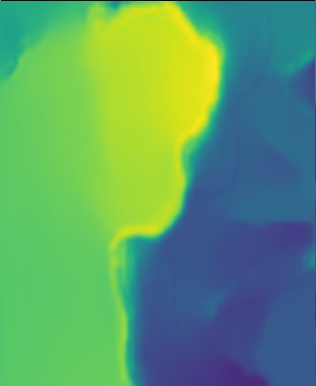}} \\
    & & \vspace{-18.8pt}\hspace{-10pt}\includegraphics[width=0.8\linewidth]{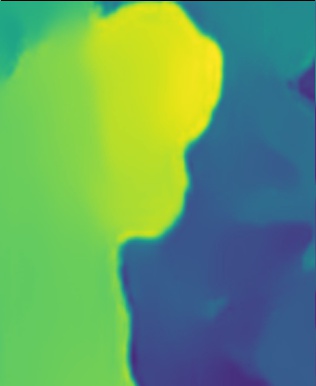}
    \end{tabular}
    \caption{Comparison of (top row) inpainted intensity and (bottom row) depth images with and without being trained on \emph{KeystoneDepth}. Left: Input image. Middle: Inpainted results with the complete training pipeline. Right: Close-ups of the results. The first and third rows show the results after pre-training on synthetic SUNCG data, and the second and fourth rows are the result of our full training sequence that includes \emph{KeystoneDepth}.
    }
    \label{fig:finetune}
\end{figure}

We also evaluated the benefit of using data from the \emph{KeystoneDepth} dataset to train our inpainting networks. As illustrated in Figure~\ref{fig:finetune}, depth edges become noticeably sharper and interior details in the intensity images are better preserved when our training includes the \emph{KeystoneDepth} data as opposed to stopping after the pre-training phase that uses only synthetic SUNCG data. We attribute this to the fact that the \emph{KeystoneDepth} data is naturally more representative of this particular task and so it does a better job of guiding the learning process.

\paragraph{View extrapolation:}
We compared our method to Stereo Magnification~\cite{zhou2018stereo}, which learns a multi-plane image (MPI) representation from a stereo pair that allows synthesizing novel views. We believe this method is the most similar to our approach as other techniques either require more than two input views~\cite{hedman18,penner2017soft3dreconstruction} or are not compatible with real-time applications on low-end devices~\cite{choi2018extremenvs}.

\begin{figure}
\centering
\setlength{\tabcolsep}{3pt}
\def\imw{0.42\linewidth}
\begin{tabular}{cccc}
    \includegraphics[trim={0 2cm 0 0},clip,width=\imw]{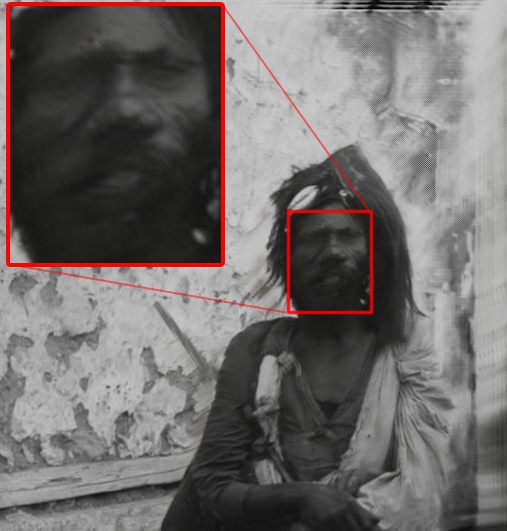} & 
    \includegraphics[trim={0 2cm 0 0},clip,width=\imw]{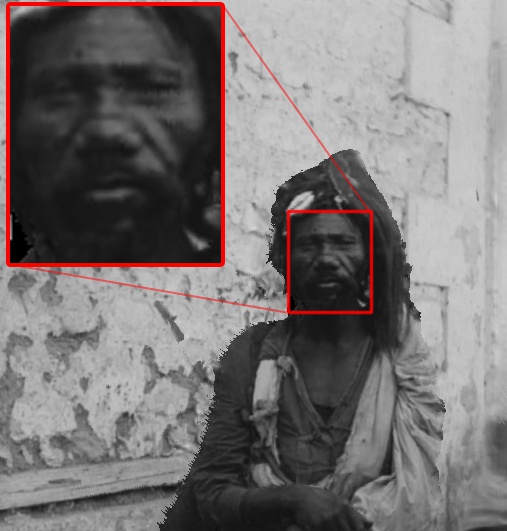} \\
    \includegraphics[width=\imw]{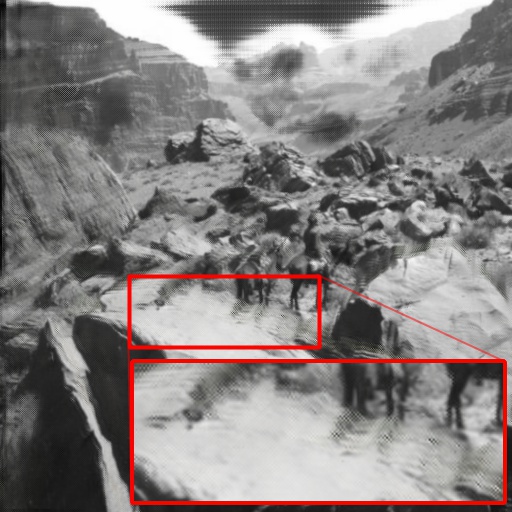} & 
    \includegraphics[width=\imw]{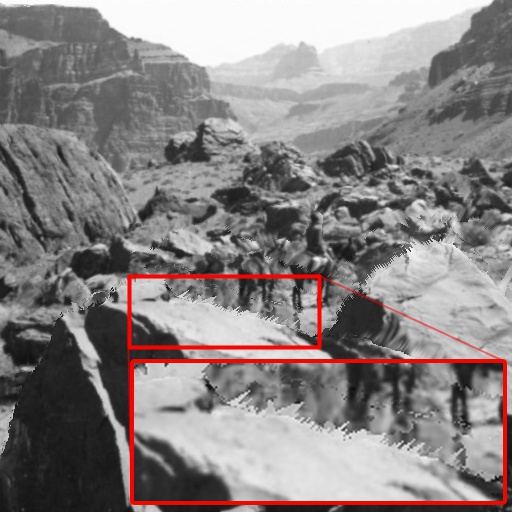}
\end{tabular}
    \caption{Comparison of view synthesis results produced by (left) Stereo Magnification~\cite{zhou2018stereo} and (right) our method.}
    \label{fig:compare-stereo-magnification}
\end{figure}

Figure~\ref{fig:compare-stereo-magnification} shows a comparison to Stereo Magnification using their publicly released code.
Note that unlike our technique, theirs was not trained using \emph{KeystoneDepth} data as it requires videos of panning motion.
We found that the most noticeable artifact produced by Stereo Magnification for this application is the so called "stack of cards" effect~\cite{zhou2018stereo}, due to it struggling to locate foreground and background elements onto separate planes. Our approach, on the other hand, is able to support a larger viewing volume due to it maintaining an explicit model of the scene geometry and inpainting underlying holes in the background.
We also observed depth quantization artifacts in the MPI representation, most noticeably near slanted surfaces that straddle multiple layers in the stacked representation. Overall, we believe our approach provides better visual performance for this particular dataset and application. Our supplemental materials includes additional comparisons.

We also performed a blind user study to compare our approach to Stereo Magnification~\cite{zhou2018stereo}. This study presented 20 randomly selected stereograph scenes to test subjects (N=29), where for each technique we showed a short 7 second virtual camera path exhibiting parallax. We asked the  participants to rate both algorithms for each test stereograph on a 1 to 7 Likert scale where 7 corresponds to ``A is much better than B'', 6 to ``better'', 5 to ``a bit better'', 4 to ``equally good'', etc. We found that $62.3\%$ of the subjects preferred our approach, $27.8\%$ preferred Stereo Magnification, and the balance had no preference.

\section{{\em KeystoneAR} Mobile App}
Finally, we developed an Augmented Reality (AR) mobile application that simulates the experience of looking through a window into a historical scene captured in these antique stereographs. It works by detecting the plane of a wall facing the user and then places a simulated window on the wall and uses our view synthesis technique to simulate a 3D historical scene behind the window. The user can move around and peer into the scene through the window, as shown in Figure~\ref{fig:teaser}d. Please see the demonstration video in the supplementary materials.
\section{Conclusion and Future Work}

This paper introduces one of the world's largest and most diverse collections of rectified stereo images and depthmaps, \emph{KeystoneDepth}, which captures historical people, events, objects and scenes between 1860 and 1963. We believe this dataset will inspire and support future research into bringing these lost moments to life. We also described a novel novel lightweight 3D scene representation and inpainting technique for intensity and depth images that allows re-rendering novel viewpoints within a viewing volume centered around these input stereo images. Our inpainting technique uses a neural network that is trained using samples extracted from the \emph{KeystoneDepth} data, using a new double reprojection method. We also incorporate a depth boundary guidance input image to the network that helps reproduce sharp and consistent depth edges. Finally, we integrated all of these components into a mobile AR application that gives the sensation of looking through an open window into these historical scenes.

There are many avenues for future work.  While we produce good results for thousands of scenes, others fail due to failures in calibration, rectification, or disparity estimation. Our inpainting network is far from perfect and our scene representation does not guarantee hole-free interpolations, particularly in scenes with complex occlusions. Other interesting directions include improving image quality, hallucinating motion (or sound) in these scenes, and expanding the effective camera field of view.


\section{Acknowledgements}
We thank Ying Wang for help on the dataset, Ayush Sarah and Brennan Stein for help on the KeystoneAR App. We are grateful to Konstantinos Rematas, Alex Colburn, Kiana Ehsani and Aditya Sankar for many helpful comments. We thank Rick Szeliski for introducing us to Slow Glass. This work was supported by the UW Reality Lab, Animation Research Lab, Facebook, Google, and Huawei. 

{\small
\bibliographystyle{ieee}
\bibliography{egbib}
}

\end{document}